\documentclass[acmtog]{acmart}
\setcitestyle{square}

\usepackage{booktabs} % For formal tables
\usepackage{float}
\usepackage[ruled]{algorithm2e} % For algorithms
\usepackage[toc,page]{appendix}
\usepackage{epigraph}
\usepackage[T1]{fontenc}
\usepackage{lmodern}

\begin{document}
\setcopyright{acmlicensed}
\acmJournal{TOG}
\acmYear{2018}\acmVolume{37}\acmNumber{6}\acmArticle{1}\acmMonth{11}
\acmDOI{10.1145/3272127.3275020}

\title{Video to Fully Automatic 3D Hair Model}

\author{Shu Liang}
\affiliation{%
  \institution{University of Washington}
}
\email{liangshu@cs.washington.edu}

\author{Xiufeng Huang}
\affiliation{%
  \institution{Owlii}
}
\email{xiufeng.huang@owlii.com}

\author{Xianyu Meng}
\affiliation{%
  \institution{Owlii}
}
\email{xianyu.meng@owlii.com}

\author{Kunyao Chen}
\affiliation{%
  \institution{Owlii}
}
\email{kunyao.chen@owlii.com}

\author{Linda G. Shapiro}
\affiliation{%
  \institution{University of Washington}
}
\email{shapiro@cs.washington.edu}

\author{Ira Kemelmacher-Shlizerman}
\affiliation{%
  \institution{University of Washington}
}
\email{kemelmi@cs.washington.edu}

\renewcommand{\shortauthors}{S. Liang et al.}

\begin{abstract}

Imagine taking a selfie video with your mobile phone and getting as output a 3D model of your head (face and 3D hair strands) that can be later used in VR, AR, and any other domain. State of the art hair reconstruction methods allow either a single photo (thus compromising 3D quality) or  multiple views, but they require  manual user interaction (manual hair segmentation and capture of fixed camera views that span full $360
^{\circ}$).  In this paper, we describe a system that can completely automatically create a reconstruction from any video (even a selfie video), and we don't require specific views, since taking your $-90^{\circ}$, $90^{\circ}$, and full back views is not feasible in a selfie capture.  

In the core of our system, in addition to  the automatization components, hair strands are estimated and deformed in 3D (rather than 2D as in state of the art) thus enabling superior results. We provide qualitative, quantitative, and Mechanical Turk human studies that support the proposed system, and show results on a diverse variety of videos (8 different celebrity videos, 9 selfie mobile videos, spanning age, gender, hair length, type, and styling).

%Given a video of a person's head, our method automatically reconstructs a full 3D head model. State of the art method for hair reconstruction assumed  selection of four views of a person's head and manual hair segmentation. In this paper, we present an algorithm  that  uses uncalibrated views from a video instead of four calibrated views and  achieves superior results. Unlike state of the art, the presented algorithm does not require user intervention, thus can be applied on many diverse videos as demonstrated in the experiments on variety of selfie videos and celebrity online videos. 

\end{abstract}

\begin{CCSXML}
<ccs2012>
<concept>
<concept_id>10010147.10010371.10010396</concept_id>
<concept_desc>Computing methodologies~Shape modeling</concept_desc>
<concept_significance>500</concept_significance>
</concept>
</ccs2012>
\end{CCSXML}

\ccsdesc[500]{Computing methodologies~Shape modeling}

\keywords{Hair Reconstruction, Selfie video, 3D hair, in the wild}

\begin{teaserfigure}
  \includegraphics[width=\textwidth]{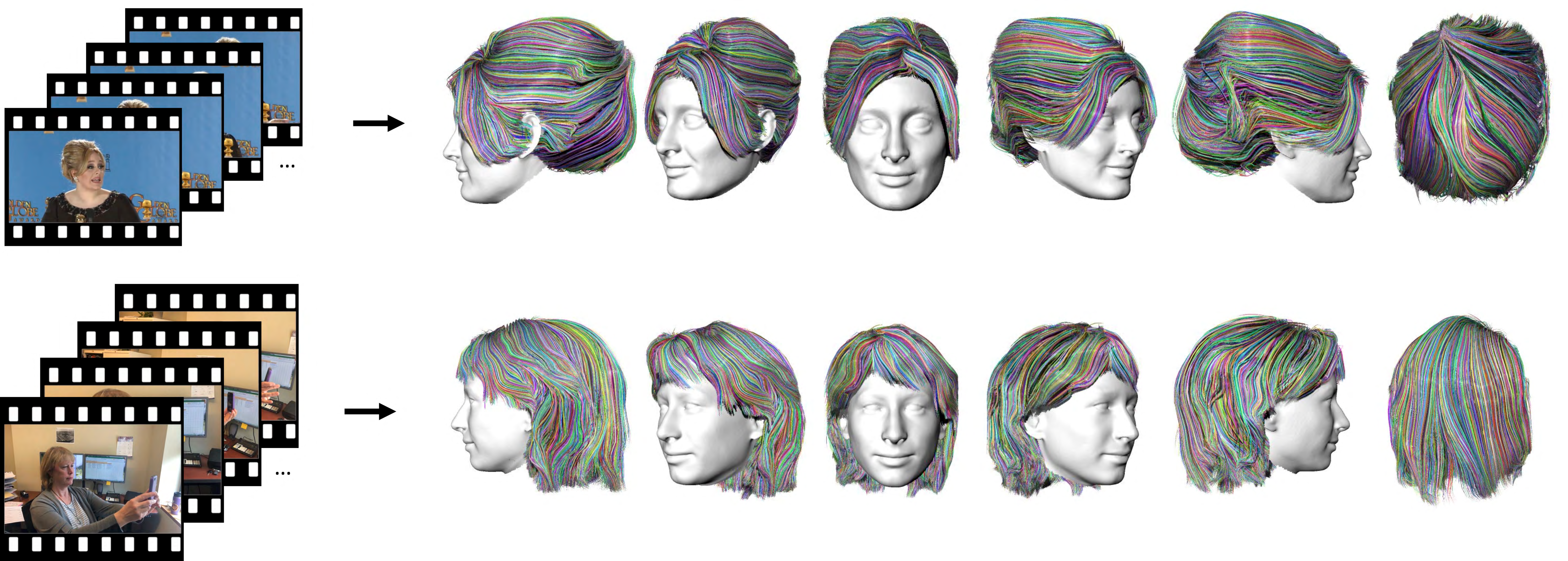}
  \caption{Two example video inputs (left)   and automatic reconstructions produced by our method (right). Input to the algorithm is any video that captures a person's head, e.g., in the figure we show a celebrity video and a selfie video captured by a mobile phone (figure shows observer viewpoint). }
  \label{fig:teaser}
\end{teaserfigure}

\maketitle

\section{Introduction}
\label{introduction}
\begin{figure*}
  \includegraphics[width=\textwidth]{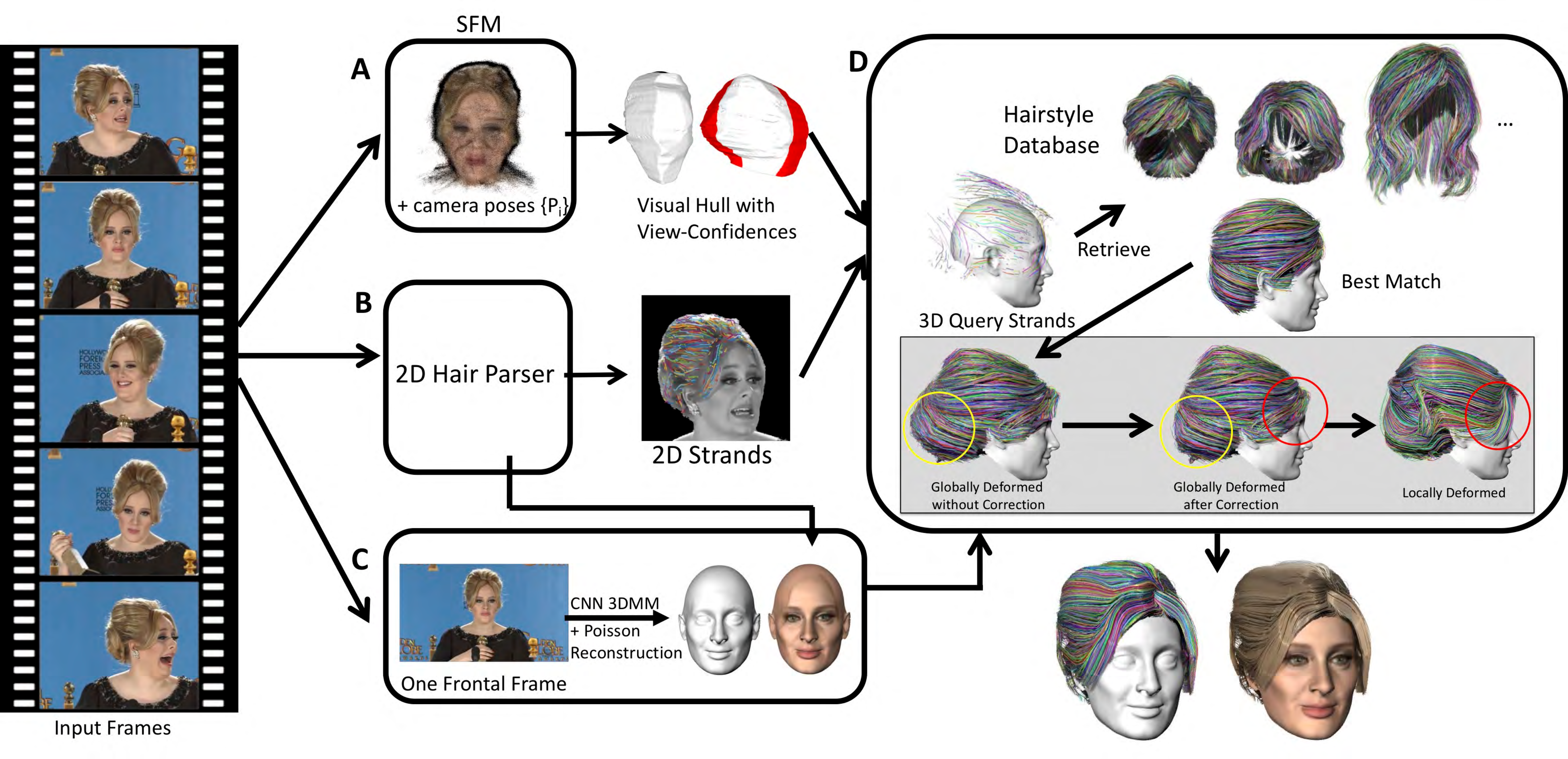}
  \caption{Overview of our method. The input to the algorithm is a video: (A) structure from motion is applied to the video to get camera poses, depth maps and a visual hull shape with view-confidence values, (B) hair segmentation and gradient direction networks are trained to apply on each frame and recover 2D strands, (C) the segmentations are used to recover the texture of the face area, and a 3D face morphable model is used to estimate face and bald head shapes. The core of the algorithm is (D) where the depth maps and 2D strands are used to create 3D strands, which are used to query a hair database; the strands of the best match are refined globally and locally to fit the input photos.  }
  \label{fig:pipeline}
\end{figure*}

\epigraph{Her glossy hair is so well resolved that
Hiro can see individual strands refracting the light into tiny
rainbows.}{\textit{Snow Crash, Neal Stephenson}}

Any future virtual and augmented reality application that includes people must have a robust and fast system to capture a person's head (3D hair strands and face). The simplest  capture scenario is  taking a single selfie photo with a cell phone \cite{Hu:2017:ADS:3130800.31310887}. 3D reconstruction from a single selfie \cite{chai2016autohair}, however, by definition, will not produce high fidelity results due to the ill-posedness of the problem--a single view does not show sides  of the person. Using multiple frames, however, will create an accurate reconstruction.  Indeed, a state-of-the-art method for hair modeling \cite{zhang2017data} needs four views, but it requires spanning the  full  $360
^{\circ}$ (front, back, and sides), as well as user interaction. This paper proposes to use a video as input and introduces solutions to three  obstacles that prevent state-of-the-art work \cite{zhang2017data,chai2016autohair,Hu:2017:ADS:3130800.31310887} from being applicable in simple automatic self capture:
\begin{enumerate}
	\item Fixed views: \cite{zhang2017data} requires four views (front, back, and two side views), those are hard to acquire accurately with self capture. In this paper, we do not constrain the views; instead we use any available video frames in which the subjects talk or capture themselves. For the input videos which do not have a full view range from $-90$ to $90$ degrees, we will correct the incomplete views with a hair shape from a database. Camera poses are estimated automatically with structure from motion. Results with various view ranges are demonstrated (as low as 90 degrees range). \cite{chai2016autohair,Hu:2017:ADS:3130800.31310887} assume a single frontal image.  
	
	\item Hair segmentation: \cite{zhang2017data} relies on the user to label hair and face pixels. In this paper, we use automatic hair segmentation and don't require any user input. Using general video frames, rather than four fixed views, introduces motion blur, varying lighting, and resolution issues, all of which our system overcomes. 
	
	\item Accuracy: our method compares and deforms hair strands in 3D rather than 2D, and the availability of the back view is not required as in \cite{zhang2017data}. It achieves higher accuracy results as demonstrated with qualitative, quantitative, and human studies. Intersection of union rate of the hair region compared to ground truth photos is on average 80\% for our method (compared to 60\% by \cite{zhang2017data}). Amazon Turk raters prefer our results 72.4\% over the four-view method \cite{zhang2017data} and 90.8\% over the single-view method of \cite{Hu:2017:ADS:3130800.31310887}.
	
\end{enumerate}

In addition to \cite{zhang2017data} (and a previously multi-view based method by  \cite{vanakittistien20163d} that also required user interaction), there is a large body of work for modeling hair from photos. Earlier works assumed laboratory calibrated photos, e.g.,  \cite{hu2014robust}. More recently \cite{chai2013dynamic,hu2015single,Hu:2017:ADS:3130800.31310887,chai2016autohair} showed how to reconstruct hair from a single photo. Interesting hair-related applications inspire further research, e.g., depth-based portraits \cite{chai2016autohair}, effective avatars for games \cite{Hu:2017:ADS:3130800.31310887}, photo-based hair morphing \cite{weng2013hair}, and hair-try-outs \cite{kemelmacher2016transfiguring}. Enabling reconstruction ``in the wild'' is an open problem where Internet photos have been explored  \cite{liang2016head}, as well as structure from motion  \cite{ichim2015dynamic} on a mobile video input. Both methods output a rough structure of the hair and head, without hair strands. This paper proposes a system that can  take in an in-the-wild video and automatically output a full head model with a 3D hair-strand model.

\section{Related Work}
\label{related}

In this section, we describe work that focuses on face, head, and hair modeling. 

\textbf{Face modeling} (no hair, or head) has progressed tremendously  in the last decade.  Beginning with high-detailed head geometry with a stereo capturing system \cite{beeler2010high,debevec2012light,alexander2013digital}, then RGB-D-based methods like dynamic fusion \cite{newcombe2015dynamicfusion} and non-rigid reconstruction methods \cite{thies2015real,zollhofer2014real} allowed capture to be real-time and much easier with off-the-shelf devices. \cite{blanz1999morphable} proposed a 3D morphable face model to represent any person's face shape using a linear combination of face bases,  \cite{tran2017regressing,richardson20163d,richardson2017learning} proposed CNN-based systems and \cite{kemelmacher2011face,suwajanakorn2014total,kemelmacher20113d,suwajanakorn2015makes}  showed how to estimate highly detailed shapes from Internet photos and videos.

\textbf{Head modeling:} Towards creating a full head model from a single photo,  \cite{chai2015high}  modeled the face with hair as  a 2.5D portrait, using head shape priors and shading information.  \cite{maninchedda2016semantic} allowed multiple views and used volumetric shape priors to reconstruct the geometry of a human head starting from structure-from-motion dense stereo matching.   \cite{cao2016real} showed that a full 3D head with a rough but morphable hair model can be reconstructed from a set of captured images  with hair depth estimated from each image and then fused together. Finally,  \cite{liang2016head} explored how to reconstruct a full head model of a person from Internet photos.  These methods focused on rough head and hair modeling without reconstruction of hair strands.

\textbf{Hair strand modeling} is key to digital human capture. Capturing the nuances of person's hair shape is critical, since the hair style is a unique characteristic of a person and inaccuracies can change their appearance dramatically. 

Multi-view camera rigs and a controlled capturing environment were able to acquire hair shape with high fidelity  
\cite{luo2013structure,hu2014robust,paris2008hair,paris2004capture,ward2007survey},  and more recently with RGB-D cameras \cite{hu2014capturing}. With a single RGB image, \cite{chai2012single,chai2013dynamic} showed that strands can be recovered from per pixel gradients, and with a use of a database of synthetic hairstyles, \cite{hu2015single} created a natural-looking hair model. A key requirement was to have a user draw  directional strokes on the image to initialize strand construction.  A fully automatic approach was proposed recently by \cite{chai2016autohair} and \cite{Hu:2017:ADS:3130800.31310887}  with CNN-based methods for hair segmentation, direction classification and a larger database for retrieving the best match for a single-view input. 

Since single-view modeling is ill-posed by definition,  \cite{vanakittistien20163d} used a hand-held cell phone camera to take photos from $8$ views of the head to recover the hair strands, and \cite{zhang2017data} proposed a method to reconstruct the hair from four-view images starting from a rough shape retrieved from a database and synthesized hair textures to provide hair-growing directions to create detailed strands. However, both methods need human interactions for hair segmentation and pose alignment.  This paper solves those constraints and also demonstrates higher accuracy results compared to those methods.

\section{Overview} 

% Our pipeline is shown in Fig \ref{fig:pipeline}. To get a personalized reconstruction of a subject's head, we need information for both face and hair details. However, most of the preliminary works are not capable of reconstructing a complete head of the person because they failed to use a person's photos or videos in different views. Inspired by the work \cite{liang20143d} that starts from a rough shape from a single depth frame from Kinect and fills in the details from a high-resolution 3D dataset. We start from a person's Internet video sequences and augment the result with a synthetic 3D hairstyle dataset and propose the following work:
% \begin{enumerate}
% \item Recovering the rough shape from video frames,
% \item Generating the facial part using a 3D Morphable head model,
% \item Augmenting hair details from the hairstyle dataset.
% \end{enumerate}

Figure~\ref{fig:pipeline} provides an overview of our method and the key components. The input to the algorithm is a video sequence of a person talking and moving naturally, as in a TV interview. There are four  algorithmic components (correspond to the labeling of boxes in Figure \ref{fig:pipeline}): 

(A) video frames are used to create a structure-from-motion model, estimate camera poses as well as per-frame depth, and compute a rough visual hull of the person with view-confidences, (B) two models are trained: one for hair segmentation, and another for hair direction; given those models, 2D hair strands are estimated and hair segmentation results are transferred to the visual hull to define the hair region, (C) the masks from the previous stage are used to separate the hair from the face and run the morphable model (3DMM) to estimate the face shape and later create the texture of the full head. 

(D) is a key contribution in which first  depth maps and 2D hair strands are combined to create 3D strands, and then 3D strands are used to query a database of hair styles. The match is deformed according to the visual hull, then corrected based on the region of confidence of the visual hull. Finally, it is deformed on the local strand level to fit the input strands. Texture is estimated from input frames to create the final hair model. The full head shape is a combination of the face model and the hair strands. In the next sections, we describe each of these components. 
\begin{figure}
	\includegraphics[width=0.5\textwidth]{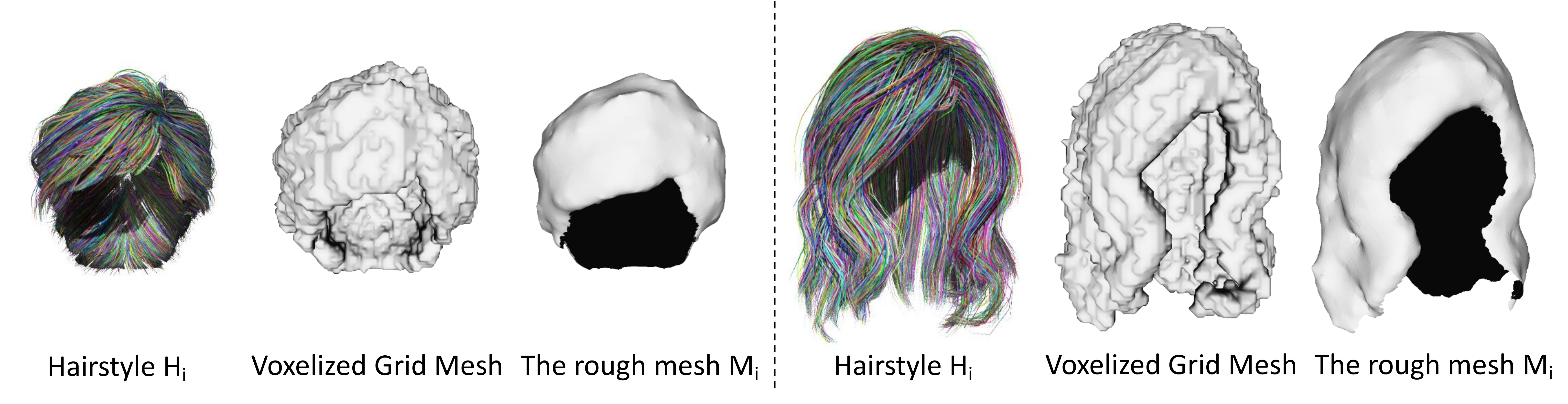}
	\caption{Example hair styles from the dataset. For each hairstyle $H_i$, we create its corresponding rough mesh $M_i$ as described in the text. }
	\label{fig:hairmesh}
\end{figure}

\begin{figure*}
	\includegraphics[width=\textwidth]{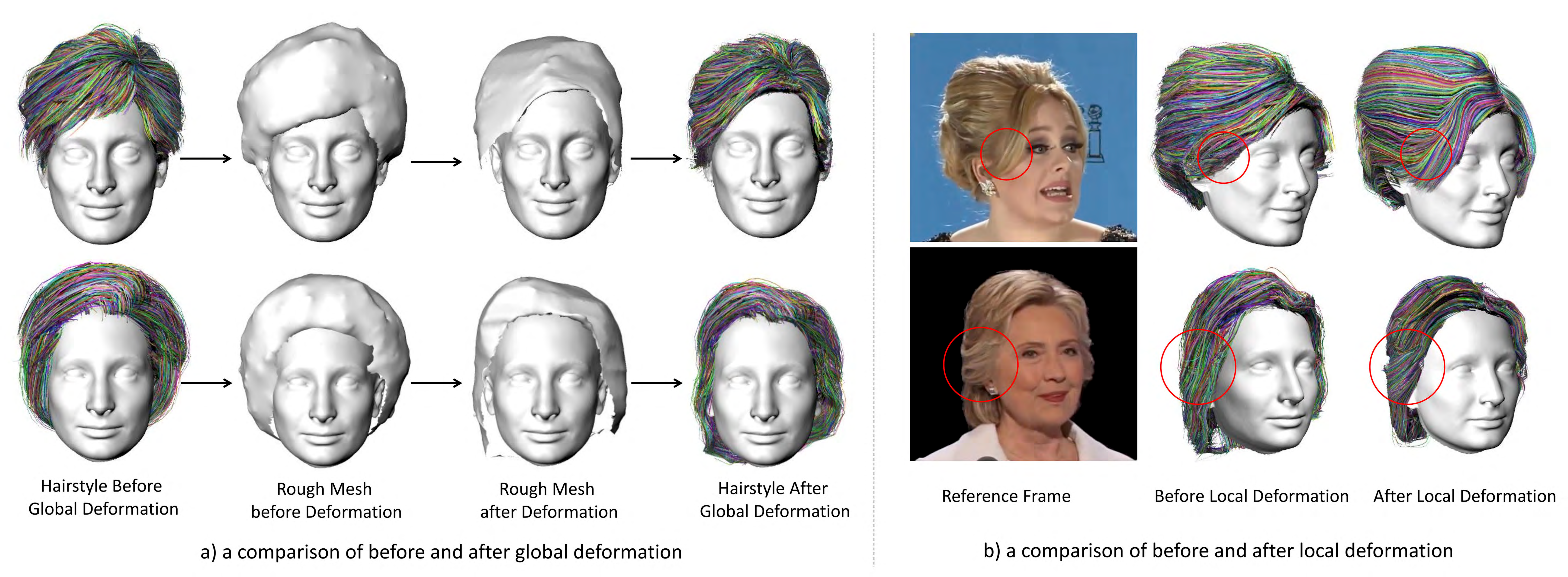}
	\caption{In (a), we show a comparison of before and after global deformation. The retrieved hairstyle is deformed under the control of its rough mesh to fit the visual hull shape. In (b), we show a comparison of before and after local deformation. A video frame is shown as a reference that after local deformation, we are able to recover more personalized hair details.}
	\label{fig:deformation}
\end{figure*}

\section{Input Frames to Rough Head Shape}\label{A}
This section describes part (A) in Figure \ref{fig:pipeline}. 
We begin by preprocessing each frame $i$  using semantic segmentation to roughly separate the person from the background \cite{zheng2015conditional} resulting in masks $S_i$. Our goal is to estimate camera pose per frame and to create a rough initial structure from all the frames. Since the background is masked out, having the head moving while the camera is  fixed is roughly equivalent to the head being fixed while the camera is moving; thus we use structure from motion \cite{wu2011visualsfm} to estimate camera pose $P_i$ per frame and per-frame depth $D_i$ using \cite{goesele2007multi}.

Given $S_i$ and $P_i$ per frame, we estimate an initial visual hull of the head using  shape-from-silhouette \cite{laurentini1994visual}. The method takes a list of pairs $P_i$ and  $S_i$ as input and carves a 3D voxel space to obtain a rough shape of the head. Meanwhile, each segmented video frame is processed using the IntraFace software \cite{XiongD13}, which provides $49$ inner facial landmarks per frame. The 2D facial landmarks are transferred to 3D using $D_i$ and averaged.

The hair segmentation classifier trained in step(B) (Section \ref{B}) is run on all of our video frames. Each pixel is assigned a probability of being in the hair region. We drop the video frames with large motion blurs by calculating the surface area $S_i$ of the detected hair region on each frame. Assuming the head size is relatively fixed across frames, a valid frame should have a hair region size of at least $0.33 \overline{S_i}$. The corresponding probabilities of the valid frames are transferred to the visual-hull shape. A vertex with a mean probability larger than $0.5$ is considered hair. Thus, we extract the hair part $X_h$ out of the visual-hull as shown in Figure \ref{fig:viewcorrection}(a), and the remaining is the skin part.

The resultant visual-hull shape is relatively rough due to the non-rigid nature of the subject's head and might be stretched due to the incomplete views. Ideally, assuming the camera distance is always larger than the size of the head, we will get a complete visual hull if our video covers a full azimuth range of $-90$ to $90$ degree. However, for in-the-wild videos, we usually cannot guarantee full coverage. We rigidly align the rough visual hull to a generic head model using 3D facial landmarks, and each camera pose $P_i$ is also transformed to a corresponding $P'_i$ based on this alignment. We connect each $P'_i$ to the center of the generic head (the origin point in our case) and calculate the azimuth angle $\gamma_i$ of each camera. The vertices on the visual hull with an azimuth angle in $[min(\gamma_i)-\pi/2, max(\gamma_i)-\pi/2]\cup[min(\gamma_i)+\pi/2, max(\gamma_i)+\pi/2]$ as illustrated in Figure \ref{fig:viewcorrection}(a) are denoted {\it high-confidence} vertices.

\begin{figure*}
	\includegraphics[width=\textwidth]{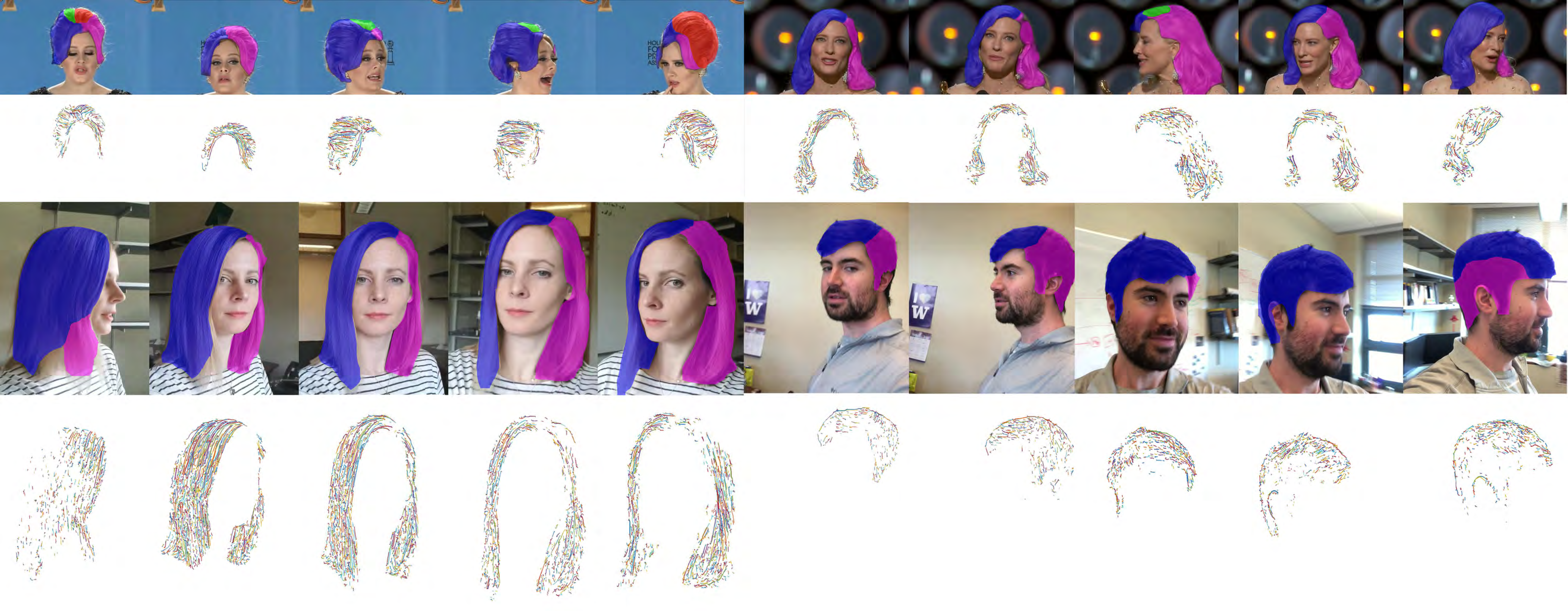}
	\caption{Examples of Figure \ref{fig:pipeline}(B). Hair segmentation, directional labels and 2D hair strands of example video frames. For the color of the directional subregions, red stands for $\left[ 0,0.5\pi\right)$, pink stands for $\left[0.5\pi,\pi\right)$, blue stands for $\left[\pi,1.5\pi\right)$ and green stands for $\left[1.5\pi,2\pi\right)$.}
	\label{fig:hairsegs}
\end{figure*}

% Our input data is a video clip of a person moving his head while making a speech. The typical frame size is around $400$ images capturing a variable range of views of his head. We first use the semantic segmentation method \cite{zheng2015conditional} to extract the person from the background. Pairwise view correspondences are computed using SIFT features \cite{lowe2004distinctive}. Since the background is masked out, the head moving while the camera is relatively fixed is equivalent to the head fixed while the camera is moving. In this case, we are able to use the structure-from-motion method \cite{wu2011visualsfm} to approximate the camera poses $P_i$ of each frame. A 2D silhouette of the head on each frame $S_i$ is obtained from the semantic segmentation result. We get an initial visual hull of the head using the shape-from-silhouette method \cite{baumgart1974geometric}. The method takes the list of pose $P_i$ and frame $S_i$ as input and carves the 3D voxel space to obtain a rough shape of the head. Theoretically, the quality of the visual-hull shape can be improved if we use more frames for carving. However, due to the pose alignment error and the non-rigid nature of the subject's head, we cannot get a very detailed mesh in the non-calibrated setting. However, we can use a 3D morphable head model to improve it.

\section{Images to 2D Strands}\label{B}
This section describes part (B) in Figure \ref{fig:pipeline}.
Inspired by the strand direction estimation method of \cite{chai2016autohair}, we trained our own  hair segmentation  and  hair directional classifiers to label and predict the hair direction in hair pixels of  each video frame. We manually label hair directional labels on our data and train a classifier using the manual labels directly as groundtruth as in \cite{chai2016autohair}. More details on our hair segmentation method can be found  in the appendix. Results of the classifier are shown on  examples in Figure \ref{fig:hairsegs} (1st and 3rd row).

To estimate 2D strands, we select one video frame every $\pi/8$ degrees according to its camera azimuth angle ${\gamma}_i$ spanning the camera view range $[min(A_i),max(A_i)]$. Similar to previous hair orientation estimation methods \cite{jakob2009capturing,chai2012single}, we filter each image with a bank of oriented filters that are uniformly sampled in $[0,\pi)$. We choose the orientation $\theta_p$ with the maximum response for each pixel to get the 2D non-directional orientation map for each image. We further trace the hair strands on each non-directional orientation map following the method in \cite{chai2012single} as shown in Figure \ref{fig:hairsegs} (2nd and 4th rows). The hair-direction labels are used to resolve the ambiguity of each traced 2D hair strand. If half of the points in a single strand have opposite directions to their directional labels, we flip the direction of the strand.

\section{Face Model}\label{C}

This section corresponds to part (C) in Figure \ref{fig:pipeline}. Each segmented video frame from the previous stage is processed using the IntraFace software \cite{XiongD13}, which provides head pose, and  $49$ inner facial landmarks. From all the frames, the frame that is closest to frontal face is picked first (where yaw and pitch are approximately $0$), and fed to a morphable-model-based face model estimator \cite{tran2017regressing}. This method generates a linear combination of the Basel 3D face dataset \cite{blanz1999morphable} with both identity shape, expression weights and texture. Here, we only use the identity weights to generate a neutral face shape. In the future, it will be easy to add facial expressions to our method. The result of the estimation is a masked face model.

% For the views that the software failed to give a valid pose detection, we apply temporal continyity  the camera poses $\{P_i\}$ to infer the head pose and corresponding facial landmarks.

% To get a personalized and detailed face shape, we pick a frontal view frame (yaw angle closest to $0$ degree) as the input for the latest 3D Morphable Model method \cite{tran2016regressing}. The method generates a linear combination of the 3D face dataset \cite{blanz1999morphable} with both identity shape, expression weights and texture. Here, we only used the identity weights to generate a neutral face shape.

% The forehead of the generated face shape is missing due to the limitation of the original 3D face dataset. 

We complete the head shape using a generic 3D head model from the Facewarehouse dataset \cite{cao2013facewarehouse}.  We choose to use the Basel dataset instead of using the Facewarehouse dataset to fit directly, because the Facewarehouse dataset contains only about $11k$ vertices for the whole head, while the Basel dataset contains about $53k$ for just the face region in which more facial shape details are provided. We pre-define 66 3D facial landmarks (49 landmarks on the inner face and 17 landmarks on the face contour and the ears) on both the 3D face dataset used by \cite{tran2017regressing} and the generic head shape. Since all the face shapes in the 3D face dataset are in dense correspondence, we  transfer these 66 landmarks to all the output 3D faces. We then deform the generic shape towards the 3D face shape using the landmarks, following \cite{liang20143d}. We fuse the deformed generic head shape and the face shape using Poisson surface reconstruction \cite{kazhdan2013screened} and get a complete head shape.

For the texture of the head, we project the full head to the selected frontal image and extract per-vertex colors from the non-hair region of the frontal image. We complete the side-view textures by projecting the head to all the frames. For the remaining invisible region, we  assign an average skin color.

\section{3D hair strand estimation}\label{D}
This section corresponds to part (D) in Figure \ref{fig:pipeline}. By utilizing a video, we deform the hairstyles in 3D instead of 2D because we are able to take advantage of the shape information of all the frames and its content continuity to estimate the per-frame pose.

\textbf{2D to initial 3D strands:}
Each video frame $i$ has an estimation of 2D strands; those are projected to depths $D_i$ to estimate 3D strands. Large peaks of the 3D hair strands (distance to the neighboring vertex larger than $0.002$ with a reference head width of $0.2$) are removed.  A merging procedure is performed to decrease future  retrieval time (and reduce duplicate strands) as follows: for each pair of strands, if their directions are the same, the pairwise point-to-point distances of the vertices in these two 3D strands are checked, and the overlapping line segments are combined. If the directions are not the same, no merging occurs. This process iterates until around $100$ 3D strands are obtained.

\textbf{3D Strands to Query a Hair Database:} The recovered 3D strands in the previous stage are sparse and incomplete; thus we use them to query a database of hair models and adjust the retrieved matches with global and local deformations to create a full hair model. The sparseness is a result of resolution of the video frames, motion blur, quality, and coverage in views (in all of our input videos, the back of the head is not visible). While being sparse, the strands do capture the person's specific hairstyle. We describe the algorithm below.  

We use the hair dataset created by \cite{chai2016autohair}, which contains $35,000$ different hairstyles, each hairstyle model consisting of more than $10,000$  hair strands. For each database hair model, we create a voxel grid around each hair strand vertex and combine all voxel grids into a voxelized mesh. The shape is further smoothed using Laplacian mesh smoothing  \cite{sorkine2004laplacian}. In order to remove the inner layer of the shape, a ray is shot from each vertex with the direction equivalent to the one from the center of the head to the current vertex. If the ray intersects any other part of the rough shape, the vertex is removed, because it is in the inner surface; otherwise it is kept. The resulting shape has   $5,000$ to $7,000$ vertices. The final cleaned shape $M$ (shown in Figure \ref{fig:hairmesh}) will be used for retrieval and deformation.

For each 3D hair strand in our query hairstyle $Q$,  the closest 3D hair strand from a hair style $H$ is determined using the following distance:
\begin{equation}\label{distance} 
Distance(Q,H)=\sum_{s_i \in Q}{\sum_{p(s_i)}{\min_{s_j \in H, \mathbf{n_{p(s_i)}} \cdot \mathbf{n_{p(s_j)}}>0}{|p(s_i)-p(s_j)|}}},
\end{equation}
where $s_i$ is a hair strand of $Q$ and $p(s_i)$ is a vertex in strand $s_i$ of $H$, $\mathbf{n_{p(s_i)}}$ is the tangent vector direction at $p(s_i)$.

This point-to-line distance comparison is very time-consuming. We performed experiments to accelerate the retrieval speed by pruning using a rough mesh $X_h$ of the head obtained from step (A) (Section \ref{A}) and step (B) (Section \ref{B}) as follows:
\begin{enumerate}
	\item{\textbf{Hairstyle boundary:}} Only the hairstyles with $x$-range in the range of ($0.8 (maxX\{X_h\}-minX\{X_h\})$, \\ $1.2(maxX\{X_h\} -minX\{X_h\})$) and $y$-range in the range of ($0.8(maxY\{X_h\}-minY\{X_h\})$,$1.2(maxY\{X_h\}-minY\{X_h\})$) are considered.
	\item{\textbf{Area of the hairstyle:}} The surface area of the rough mesh $X_h$ and of each hairstyle mesh $M_i$ are computed. Only the hairstyles with surface area in the range of ($0.8S_{X_h}$,$1.5S_{X_h}$) are considered.
\end{enumerate}

Next, the retrieved matches are deformed in global and local fashion in 3D instead of 2D, taking advantage of the multi-view information in the video. Figure \ref{fig:deformation} illustrates the deformation process. 

\begin{figure*}
	\includegraphics[width=\textwidth]{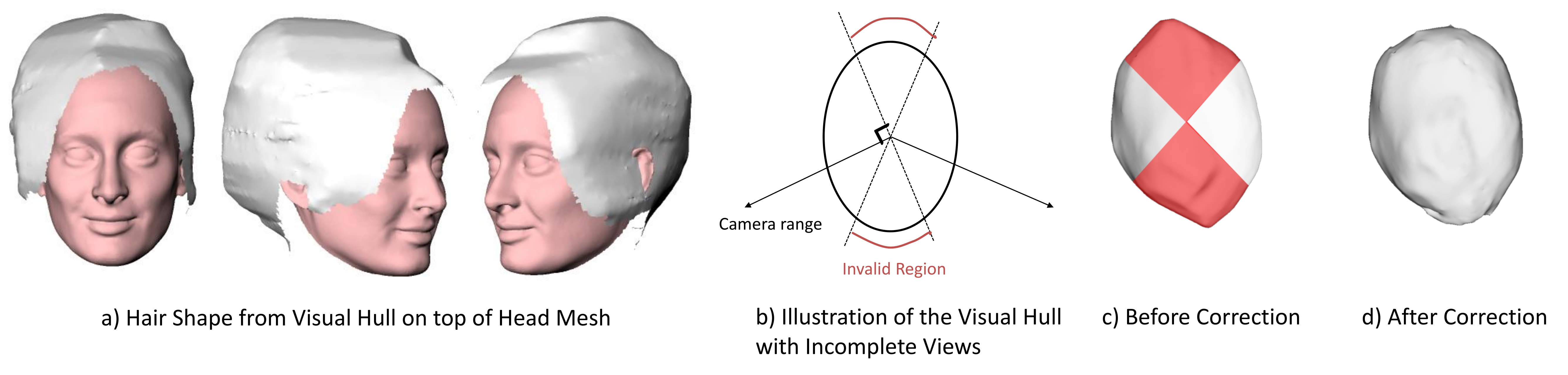}
	\caption{In (a), we show the hair part shape $X_h$ extracted from the visual hull (Figure \ref{fig:pipeline}(A)) plus the hair labels in Figure \ref{fig:pipeline}(B) on top of the head mesh from Figure \ref{fig:pipeline}(C). (b) shows an illustration of the top-down view of the visual hull with camera range and invalid regions. In (c)(d), we show a candidate hairstyle mesh $M_i$ before and after correction.}
	\label{fig:viewcorrection}
\end{figure*}

\textbf{View Correction and Global Deformation} After the top $20$ best matching hairstyles are found,  each retrieved hairstyle $M_i$ is deformed towards the rough shape $X_h$ (created by Step (A)(B) and shown in Figure ~\ref{fig:viewcorrection}(a)) using deformable registration \cite{allen} producing deformed hairstyle mesh $M_i'$. Furthermore, step (A) defines regions of {\it low-confidence} of $X_h$ (see Section \ref{A}), so further correction is needed on the corresponding regions of $M_i'$. The azimuth angle of each vertex is calculated on $M_i'$, and the vertices that are outside the confident region are considered invalid as shown in Figure \ref{fig:viewcorrection}(c) marked as red. Naturally, we think of using the original shape of $M_i$ to correct the invalid region. The correction is based on the idea of Laplacian Mesh Editing \cite{sorkine2004laplacian}. We denote the valid view range as $[R_1,R_2]\cup[R_3,R_4]$ for simplification. We assign a confidence value $c_i$ to each vertex $v_i'$ on $M_i'$ and minimize the following energy function.
\begin{equation} 
E(v_i')=\sum_{i=1}^{n}(1-c_i)||\mathcal{L}(v_i')-\mathcal{L}(v_i)||^2+\lambda\sum_{c_i=1}||v_i'-x_i||^2
\end{equation}
where $\mathcal{L}$ is a Laplacian operator, $v_i$ is the vertex position before deformation, $x_i$ is the closest point of $v_i'$ on $X_h$ after direct deformable registration, $\lambda$ is $10^{-5}$. The confidence value $c_i$ is $1$ for the valid region and is defined for the invalid region as follows:
$$ c_i=exp(-\frac{||\gamma(v_i')-R_j||^2}{2{\sigma}^2})\left\{
\begin{aligned}
&j=1, \gamma(v_i')\in (-\pi,R_1)\\
&j=2, \gamma(v_i')\in (R_2,0) \\
&j=3, \gamma(v_i')\in [0,R_3)\\
&j=4, \gamma(v_i')\in (R_4,\pi]\\
\end{aligned}
\right.
$$
where $\sigma = \pi/18$.
As shown in Figure \ref{fig:viewcorrection}(c), the stretched red region of $M_i'$ is corrected to have a natural look in (d). After the correction, a transformation matrix $T$ is obtained for each vertex on $M_i$. 

We further deform the hair strands in $H_i$ as shown in Figure \ref{fig:deformation}(a). Each vertex $v_i$ in $M_i$ works as an anchor point for the hairstyle deformation. For each point $p$ in $H_i$, its deformation will be decided by a set of neighboring anchor points as 
\begin{equation}
T_p=\frac{\alpha I + \sum_{v_i\in N(p)}{w_iT_i}}{\alpha+\sum_{v_i\in N(p)}{w_i}},
\end{equation}
where $N(p)$ is the set of neighboring anchor points chosen  to be the top $10$ closest vertices of $M_i$. $I$ is an identity matrix, and $w_i$ is defined as a Gaussian function
\begin{equation}\label{weight}
w_i = exp(-\frac{||p-v_i||^2}{2{\sigma}^2} )
\end{equation}
In our experiments, we set $\alpha$ to $0.01$ and $\sigma$ to $0.015$, while the width of our reference head is $0.2$.
We deform the top $20$ best matching hairstyles, and use the same distance function as proposed in Equation~\ref{distance} to find the final best match. We show a comparison in Figure \ref{fig:deformation} (a) before and after global deformation.

\textbf{Local Deformation} To add locally personalized hair details, we follow a method similar to \cite{fu2007sketching} by converting the deformed hair strands into a 3D orientation field $V$. The orientation of the extracted hair strands from the video frames are also added to the 3D orientation field to bend the hair strands in the surface layer of the hairstyle. For each 3D query strand, we set an influence volume with a radius of $2$ around it and diffuse it to fill in its surrounding voxels. The best matching hairstyle and the query 3D hair strands all contribute to the 3D orientation field by 
\begin{equation} 
E(\mathbf{v_i})=\sum_{i \in V}{||\Delta(\mathbf{v_i})||^2}+w_{1}\sum_{i \in C}{||\mathbf{v_i}-\mathbf{c_i}||^2}+w_2\sum_{i \in Q}{||\mathbf{v_i}-\mathbf{q_i}||^2},
\end{equation}
where $\Delta$ is the discrete Laplacian operator, $C$ is the boundary constraints with the known directions $\mathbf{c_i}$ at certain voxel grids that contain 3D strands from the best-matching hairstyle, and $Q$ is the boundary constraints from query hair strands.

In our experiment, we set the 3D orientation grid size to be $80\times 80 \times 80$, $w_1$ to be $1$ and $w_2$ to be $0.1$.
We show a comparison of results with and without the local deformation in Figure \ref{fig:deformation}(b). Notice the personalized hair strands in the red circles. We avoid the artifacts of hair going inside the head by growing new hair strands out of $V$ from the scalp region of the complete head shape of Section \ref{C} with pre-defined hair root points.

\textbf{Hair Texture} The color of each hair strand vertex is averaged from all the frames, and the unseen regions are assigned by an average color of the visible regions.

\section{Experiments}
\label{experiments}
In this section we describe the parameters used and the data collection process; we show results as well as comparisons to state-of-the-art methods. 

\subsection{Data Collection and Processing of Video Sequences}
We collected $8$ video clips of celebrities by searching for key words like "Hillary speech", "Adele award" on YouTube with an HD filter. The typical resolution of our videos is $720$p ($1280 \times 720$) with video duration around $40$ seconds sampled at $10$ fps ($380$ frames for Adele, $340$ and $240$ frames for Cate Blanchett, $250$ and $350$ frames for Hillary Clinton, $500$ frames for Justin Trudeau, $390$ frames for Theresa May, $310$ frames for Angela Merkel). The camera view point is relatively fixed across all the frames, while the subject is making a speech with his/her head turning. We processed our frames at 10fps. We ran the face detection method of \cite{XiongD13} on all the frames to determine a bounding box around the head (box height varies from $200$ to $600$). Our online video sequences typically cover the frontal, left and right view of the person. The minimum view range we have is for Angela Merkel: only $-15$ to $75$ degrees. There are no back views of any person's head in the videos. 

For mobile selfie videos, 9 subjects were asked to take a selfie video of themselves from left to right and switch hands in the front using their own smart phones (video resolution varies from $720$p to $1080$p). The subjects were not required to stay rigid and could take the video at their ease. The videos were taken in arbitrary environments and the lightings were not controlled. Note that the quality of mobile selfie videos are usually worse than the online videos due to large motion blurs caused by hand moving, auto focus, and auto exposure from phone cameras, although a higher frame resolution is accessible. The selfie video is approximately $15$ seconds sampled at 20fps ($369$ frames, $277$ frames, $229$ frames, $300$ frames, $267$ frames, $376$ frames, $383$ frames, $235$ frames and $256$ frames for each individual from top to bottom of Figure \ref{fig:mobile} ).

Later, the semantic segmentation method of \cite{zheng2015conditional} was run on video frames to remove the background and foreground occlusions such as microphones. We ran VisualSFM \cite{wu2011visualsfm} on the pre-processed frames. In \cite{wu2011visualsfm}, the non-rigid face expression change might cause large distortions in the reconstructed views; thus we set radial distortion to zero. 

\textbf{Runtime} We ran our algorithm on a single PC with a 12 core i7 CPU, 16GB of memory and four NVIDIA GTX 1080 Ti graphics cards. For a typical online video, the preprocessing and structure from motion plus visual hull in Figure \ref{fig:pipeline}(A) takes $40$ minutes. Extracting 2D query strands in Figure \ref{fig:pipeline}(B) takes $3$ minutes. The head shape reconstruction and texture extraction takes $3$ minutes to run. Hair database retrieval and deformation in Figure \ref{fig:pipeline}(D) is $40$ minutes with $500$ candidates. The 3D orientation local deformation and final hair strand generation from the reconstructed head takes $2$ min.

\subsection{Results and Comparisons}

\begin{figure*}
  \includegraphics[width=\textwidth]{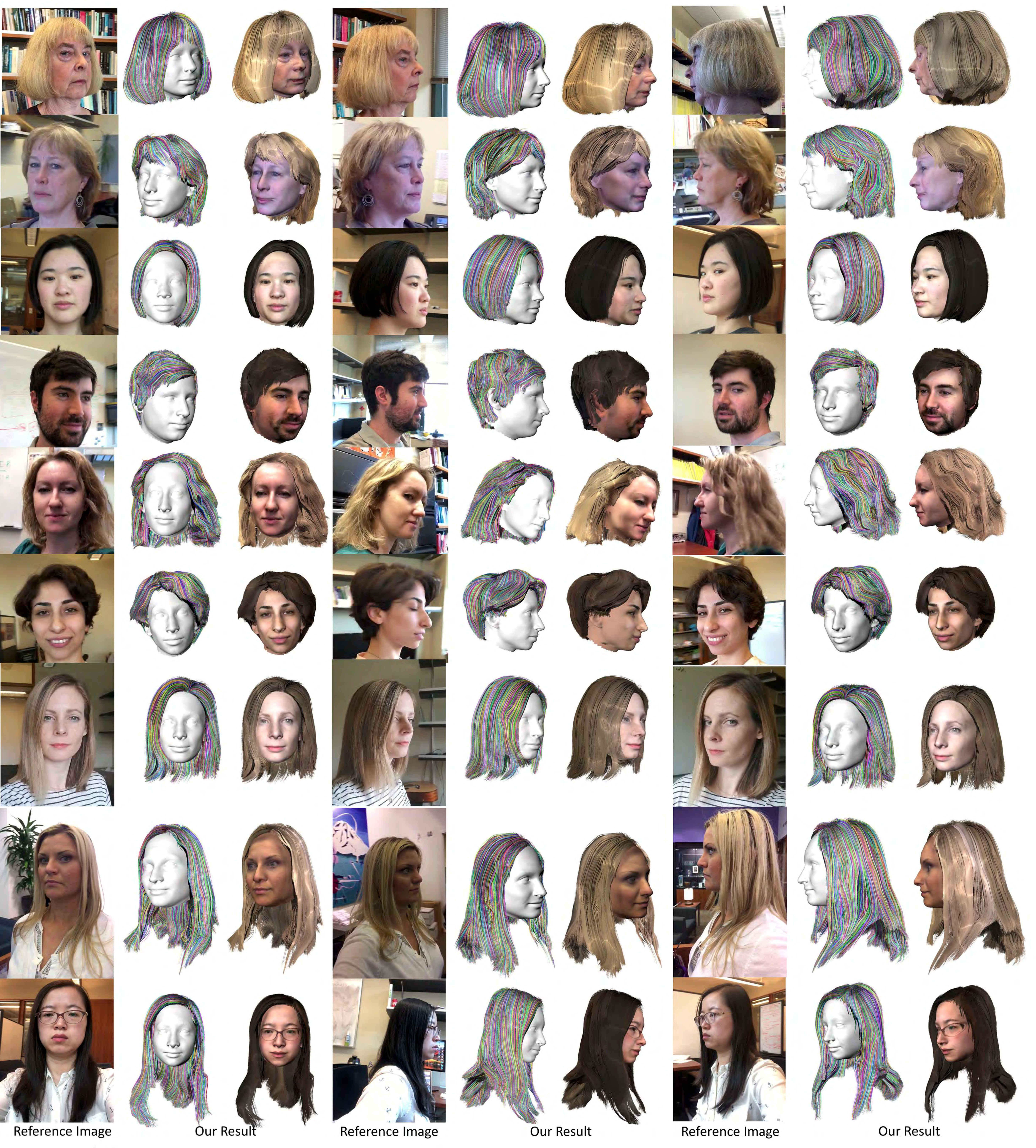}
  \caption{Reconstruction results from mobile selfie videos of different people in different environments.}
  \label{fig:mobile}
\end{figure*}

\begin{figure*}
  \includegraphics[width=\textwidth]{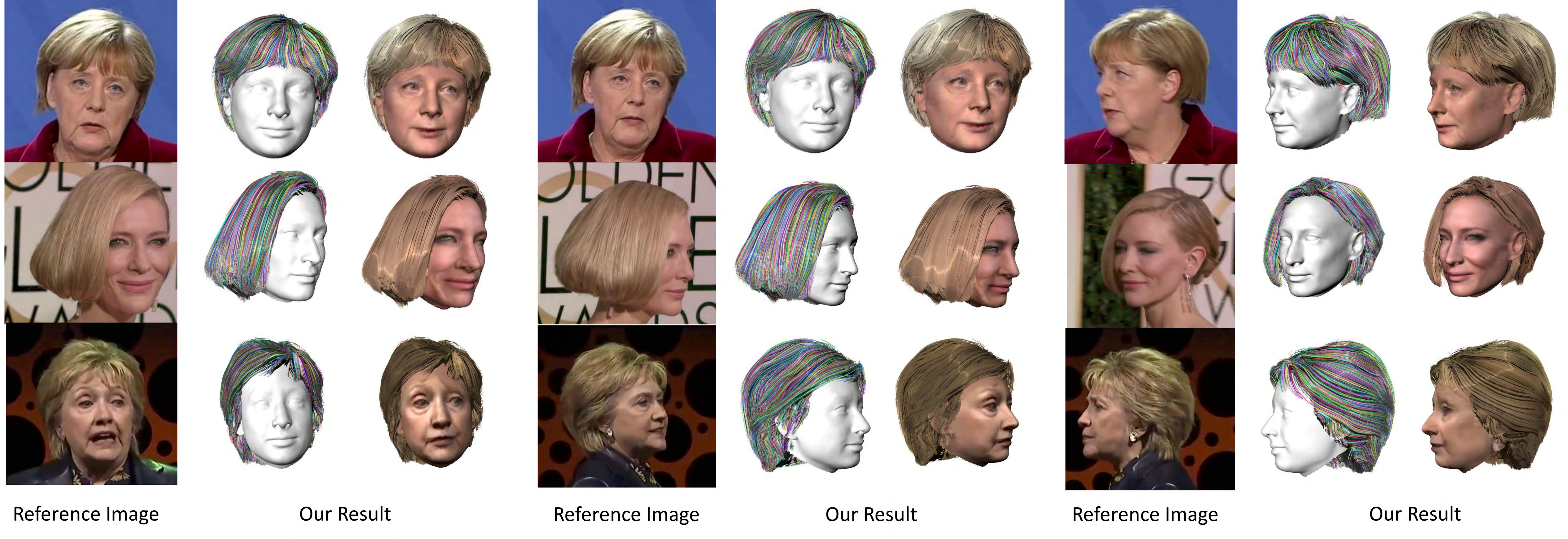}
  \caption{Example results of our method. From top to bottom, the view coverage for Angela Merkel's video is $15$ degree to $-75$ degree, $67$ to $-75$ degree for Cate Blanchett and $60$ to $-74$ degree for Hillary Clinton. Note that we can even create a natural looking result for Angela Merkel with a small view coverage.}
  \label{fig:texresult}
\end{figure*}

Figure \ref{fig:mobile} and Figure \ref{fig:texresult} show the results together with the reference frames from the videos. We can see that the reconstructions are good for a variety of lighting conditions, diverse people and hairstyles. See also the supplementary videos.

Next, we compared our results to the state-of-the-art hair in-the-wild reconstruction methods \cite{chai2016autohair,zhang2017data,Hu:2017:ADS:3130800.31310887}. More view comparisons are shown in the supplementary video.\footnote{We thank the authors of those papers for helping creating comparison results.} We performed qualitative, quantitative and user study comparisons below. 

\begin{figure*}
  \includegraphics[width=\textwidth]{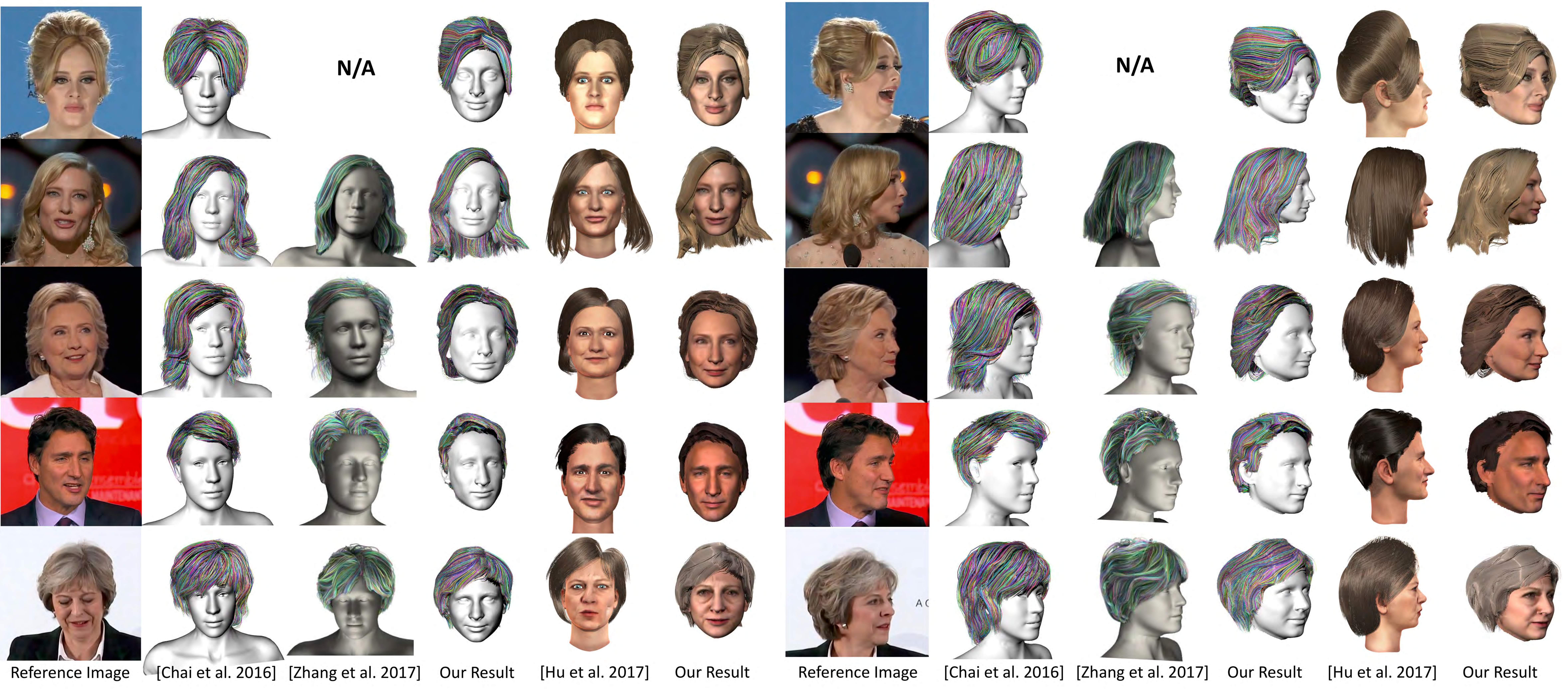}
  \caption{This figure shows our results compared to the state-of-the-art methods. For each subject, we show the results in frontal and side views. For each view, the first column shows a reference frame from the video, then we show in the order of the untextured results from \cite{chai2016autohair}, \cite{zhang2017data}, our method and the textured results from \cite{Hu:2017:ADS:3130800.31310887}, our method. Note how our result captures more personalized hair details, as also indicated by human studies and quantitative comparisons. \textbf{More view comparisons are provided in the supplementary video.}}
  \label{fig:comparison}
\end{figure*}

Figure \ref{fig:comparison} shows comparisons for single-view methods. We picked a frontal frame from each of the $5$ video clips of celebrities as input. We compared our untextured results with \cite{chai2016autohair} and textured results with \cite{Hu:2017:ADS:3130800.31310887}. Note that our 3D models captured more personalized hairstyles; for example in Adele's case (the 1st row), \cite{chai2016autohair} produced a short hairstyle, while Adele has a long hairstyle. Compared to \cite{Hu:2017:ADS:3130800.31310887}, where each hairstyle has a flat back, our results show more variety.

In \cite{zhang2017data}, frontal, left, and right views are manually chosen from the same video clip. Since we do not have the back view in our video frames and the back view is necessary for the four-view reconstruction method, the authors were allowed to use any back view image they could find to reconstruct (the authors did not reconstruct Adele; back view photos can be found in the supplementary video). In our algorithm, we did not use the back view photo of the person. Our results are similar to \cite{zhang2017data}; however ours are closer to the input; this can be seen by looking at the result of Justin (the 4th row) produced by \cite{zhang2017data} which has a larger volume.

\begin{figure*}
  \includegraphics[width=\textwidth]{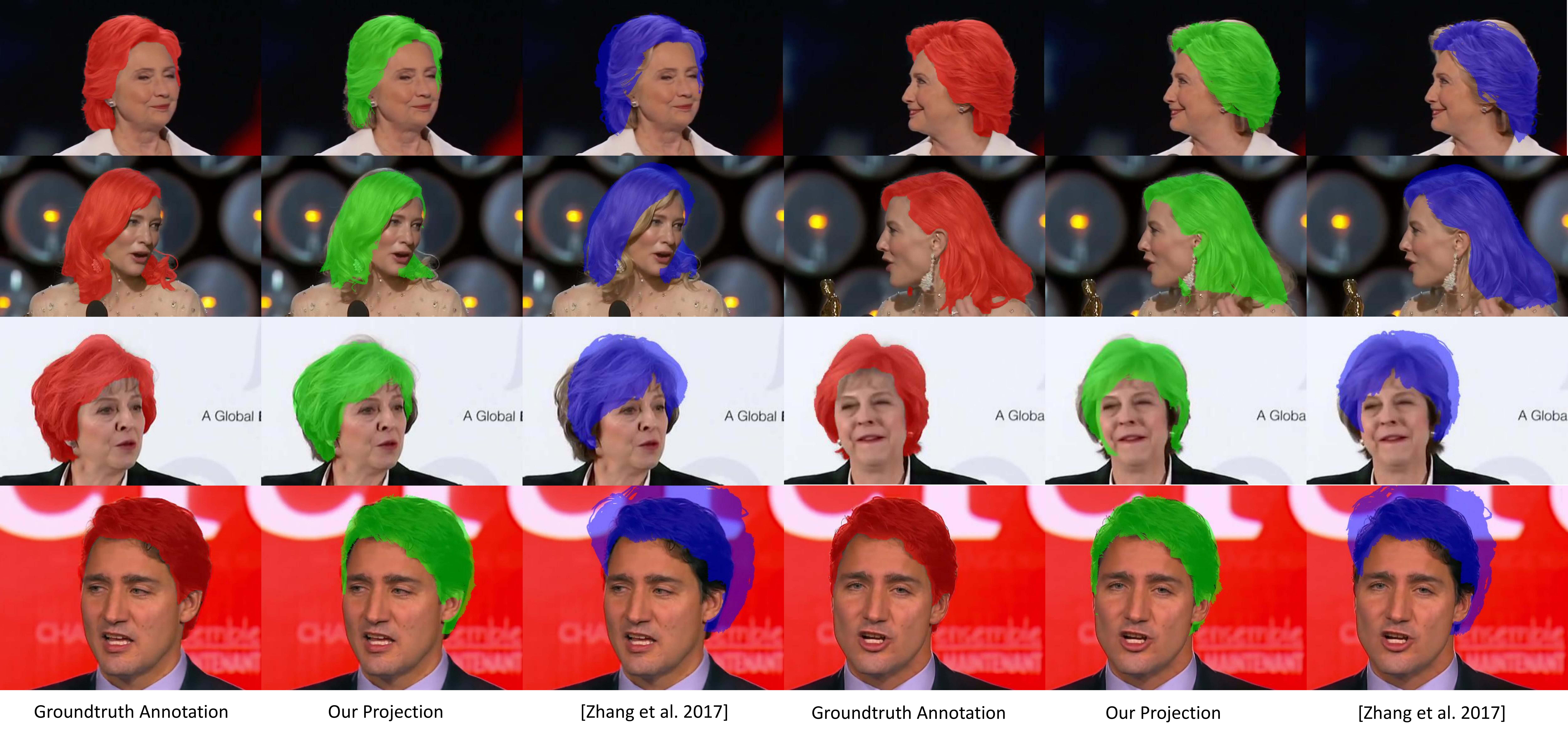}
  \caption{This figure shows four example frames comparing the silhouettes of the reconstructed hairstyles to the hair segmentation results. The red mask is the annotated groundtruth hair mask over the image frame. The green mask shows the projected silhouettes from our method over the image and the blue mask shows the projected silhouettes from \cite{zhang2017data}.}
  \label{fig:IOU2seg}
\end{figure*}

  We did a \textbf{quantitative comparison} by projecting the reconstructed hair as lines onto the images, computing the intersection-over-union rate to the ground truth hair mask (manually labeled, but not used in our training or testing of the hair classifiers) per frame. We show the average IOUs over all the frames of each subject in Table \ref{IOU2segmentation}. A larger IOU means that the reconstructed hair approximates the input better. We used the same camera pose of each frame estimated from structure from motion, where a perspective camera model is assumed, to project both the results from \cite{zhang2017data} and our results. In total, our reconstruction results get an average IOU rate of around $0.8$, while the four-view reconstruction method gets an average IOU of around $0.6$. We showed the projection and ground truth hair mask of some example frames in Figure \ref{fig:IOU2seg}. Our results resemble the hairstyles better in all the frames. Note that since the back view image used in \cite{zhang2017data} does not necessarily come from the same person, the inconsistency between four views might affect the final results. Also, in \cite{zhang2017data}, the authors assumed an orthographic camera model, which might account for some of the difference.

\begin{table}
\caption{IOU accuracy between the projected reconstructed hair and the hair segmentation (manually labeled ground truth).}
\label{IOU2segmentation}
\begin{center}
\begin{tabular}{cccc}
\hline\noalign{\smallskip}
Subject &Frames  &\cite{zhang2017data} &Ours\\
\noalign{\smallskip}
\hline 
\noalign{\smallskip}
Hillary &266 &$0.6295 \pm 0.0411$ &$\mathbf{0.8294 \pm 0.0446}$\\
Theresa &252 &$0.6598 \pm 0.0258$ &$\mathbf{0.8111 \pm 0.0216}$\\
Cate    &255 &$0.5991 \pm 0.0474$ &$\mathbf{0.7749 \pm 0.0669}$\\
Justin &307 &$0.4787 \pm 0.0882$ &$\mathbf{0.8028 \pm 0.0187}$\\
\hline 
\end{tabular}
\end{center}
\end{table}

\textbf{User Study} We performed Amazon Mechanical Turk studies to compare our results to other methods. We showed two results side by side with the ground truth image in different views and asked which shape was more similar to the input, ignoring the face, shoulder and rendering qualities. For each subject, we did $3$ groups of studies comparing the frontal, left, and right views. To remove bias, we switched the order of the two results and did $3$ more groups of studies. Each view was rated by $40$ Turkers, giving a total of $120$ different Turkers for each subject. We reported the rate of preference to our results over total in Table \ref{mturk1}. Our results achieved an average preference rate of $72.7\%$ in all the study groups. Similarly, we did a comparison of the textured results to the avatar digitalization work \cite{Hu:2017:ADS:3130800.31310887} showing the $0^{\circ}$, $15^{\circ}$ and $90^{\circ}$ views. The ratio of preferences is reported in Table ~\ref{mturk2}. In total, our results were considered better by $90.8\%$ of the Turkers.

\begin{table}
\caption{The ratio of preference to our results over total compared to \cite{zhang2017data} based on Amazon Mechanical Turk tests.}
\label{mturk1}
\begin{center}
\begin{tabular}{ccccc}
\hline\noalign{\smallskip}
Subject &frontal &left &right &total\\
\noalign{\smallskip}
\hline 
\noalign{\smallskip}
Hillary &39/40 &27/40 &27/40 &93/120\\
Theresa &13/40 &26/40 &27/40 &66/120\\
Cate    &30/40 &26/40 &32/40 &88/120\\
Justin  &27/40 &37/40 &38/40 &102/120\\
\hline 
\end{tabular}
\end{center}
\end{table}

\begin{table}
\caption{The ratio of preference to our results over total compared to \cite{Hu:2017:ADS:3130800.31310887} based on Amazon Mechanical Turk test.}
\label{mturk2}
\begin{center}
\begin{tabular}{ccccc}
\hline\noalign{\smallskip}
Subject &$0^{\circ}$ &$15^{\circ}$ &$90^{\circ}$ &total\\
\noalign{\smallskip}
\hline 
\noalign{\smallskip}
Adele   &29/40 &32/40 &38/40 &99/120\\
Hillary &35/40 &38/40 &36/40 &109/120\\
Theresa &35/40 &37/40 &39/40 &111/120\\
Cate    &40/40 &40/40 &40/40 &120/120\\
Justin  &39/40 &35/40 &32/40 &106/120\\
\hline 
\end{tabular}
\end{center}
\end{table}

\begin{figure}
  \includegraphics[width=0.5\textwidth]{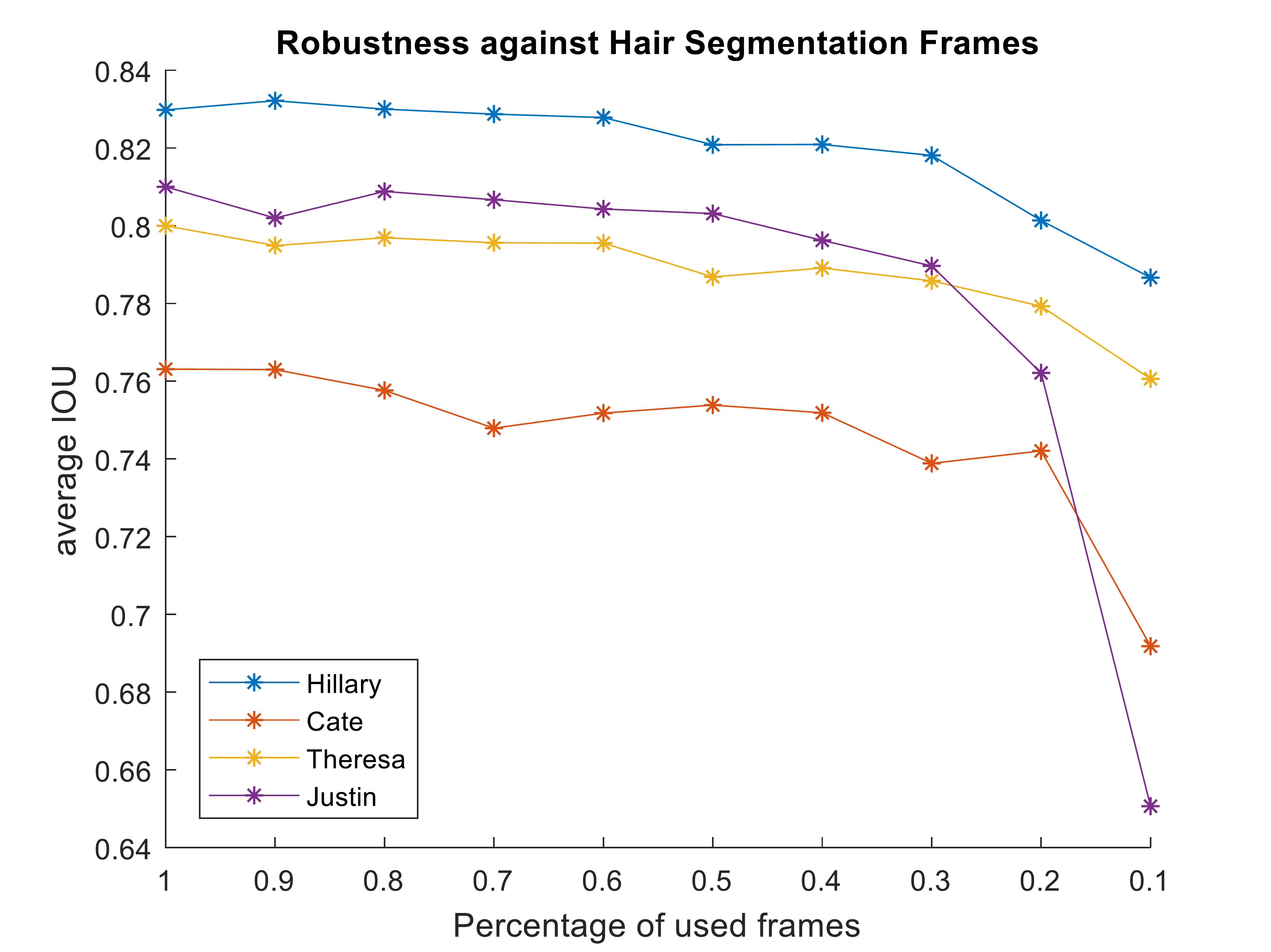}
  \caption{This figure shows how similar the rough hair shape $X_h$ to groundtruth is as the number of good hair segmentation frames decreases.}
  \label{fig:robustness}
\end{figure}

\textbf{Robustness:} To evaluate the robustness of our hair segmentation classifier, we counted the failure frames for each input video: $2$ frames for Adele, $3$ and $0$ frames for Cate Blanchette, $1$ and $20$ frames for Hillary Clinton, $0$ frames for Justin Trudeau, $0$ frames for Theresa May, and $4$ frames for Angela Merkel (2 videos each for Cate and Hillary). For selfie video inputs, the numbers are $16$ frames, $4$ frames, $3$ frames, $5$ frames, $10$ frames, $4$ frames, $3$ frames, $5$ frames and $0$ frames for each individual from top to bottom of Figure \ref{fig:mobile}. We had an average successful segmented frame rate of $98.7\%$. Generally we observed more failure segmentation frames on selfie videos due to the motion blur when the subject was switching hands and more uncontrolled lighting compared to celebrity videos. We ran simulation experiments to test how robust our system is against failure segmentation frames. We gradually decreased the number of frames (frames selected randomly) used to generate the rough hair shape $X_h$ and calculated the average IOU between the projected $X_h$ and the ground truth hair label on four celebrity videos. The result is plotted in Fig. ~\ref{fig:robustness}. Our system is quite robust even when half of the frames are dropped; however, as the number of failure frames increases we cannot guarantee a good global shape, which will lead to a poor reconstruction result.
%$380$ frames for Adele, $340$ and $240$ frames for Cate Blanchett, $250$ and $350$ frames for Hillary Clinton, $500$ frames for Justin Trudeau, $390$ frames for Theresa May, $310$ frames for Angela Merkel
%$369$ frames, $277$ frames, $229$ frames, $300$ frames, $267$ frames, $376$ frames, $383$ frames, $235$ frames and $256$ frames for each individual from top to bottom of Figure \ref{fig:mobile} 

\begin{figure}
  \includegraphics[width=0.5\textwidth]{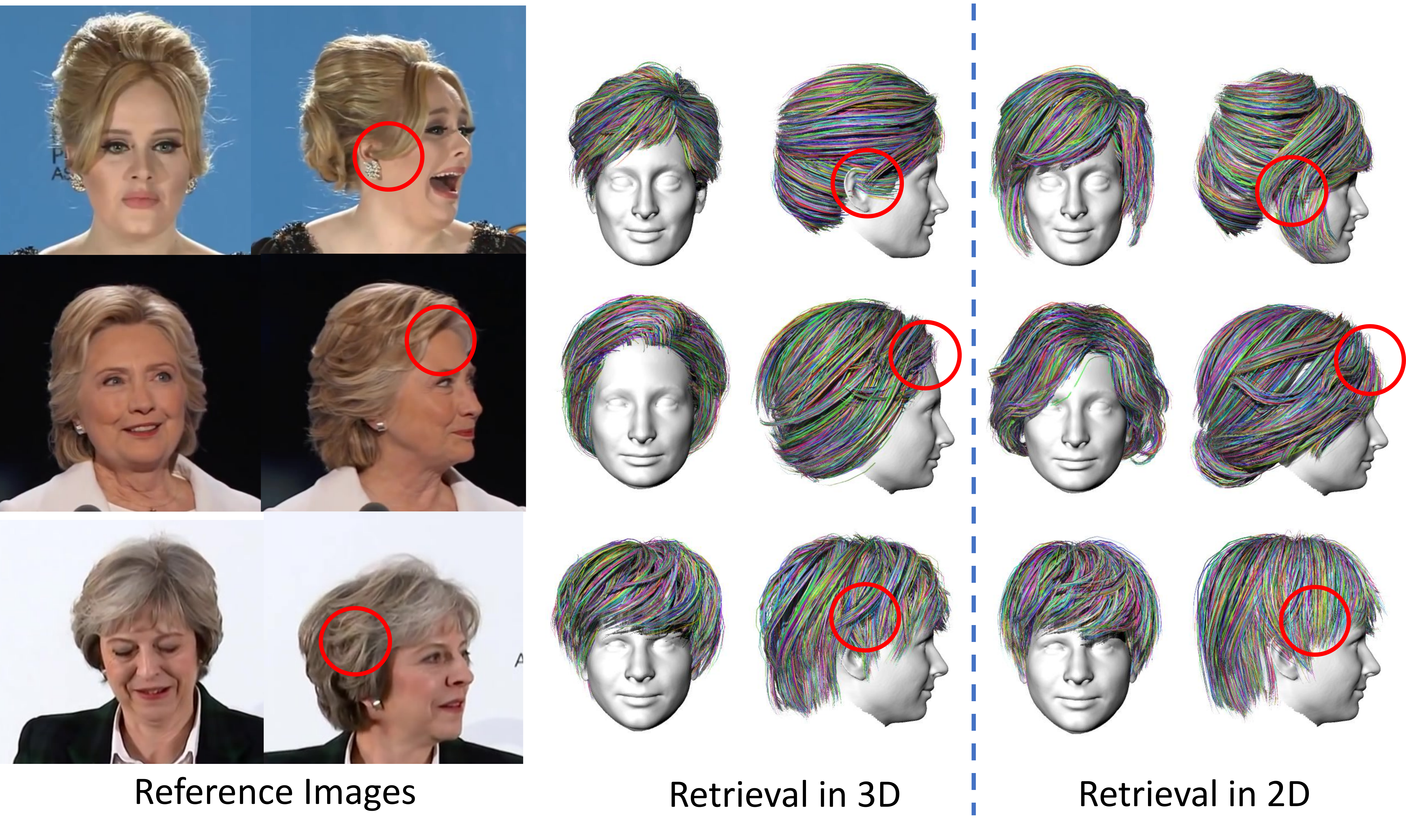}
  \caption{This figure shows the comparison for retrieving in 3D and 2D. The left two columns show two reference frames from the input video. The two columns in the middle show the frontal and side view of the best matching hairstyle from the database from 3D retrieval. The right two columns show the best matching result from 2D retrieval. The 2D retrieved results are generally similar to frontal views, however, do not look as similar as the 3D retrieval results in side views. }
  \label{fig:2dvs3d}
\end{figure}

\textbf{3D vs. 2D:} We further compared retrieving the best matching hairstyle in 3D vs retrieving in 2D. For each subject, we used the same frames as we used in 3D retrieval as query input. Instead of reprojecting each 2D strand back to 3D using per-frame depth information, we projected the candidates from the hairstyle database after pruning (Section \ref{D}) to each frame and computed the query distance similar to Equation \ref{distance}, but in 2D. We show the best matching result before any deformation in Fig. \ref{fig:2dvs3d}. 
There are three main reasons that we chose to retrieve in 3D instead of 2D. 1) The multi-frames from the same video have a lot of overlapping regions, which caused redundancy in the query input when retrieving using each 2D frame. 2) Projecting the hairstyles from the database increased the computational cost compared to projecting the sparse strands back to 3D. We might pre-generate projected 2D views to a set of pre-defined poses as in the single-view method \cite{chai2016autohair}, however, since our input video does not have fixed views, we would need to define a larger pose space than the single-view input. 3) The retrieved results are usually biased towards the frontal view, because the hair strands in relatively frontal views are seen in more frames than the side views as shown in Fig. \ref{fig:2dvs3d}. For example, in Adele's comparison, the two hair strips in the front contributed more to the retrieval, making the best matching result less like the input subject from the side view. By projecting the strands to 3D, we allow the information from different views to contribute equally to the retrieval, as well as improving computational cost. 

\section{Limitations and Applications}
\label{results}

\subsection{Limitations}
\begin{figure}
  \includegraphics[width=0.5\textwidth]{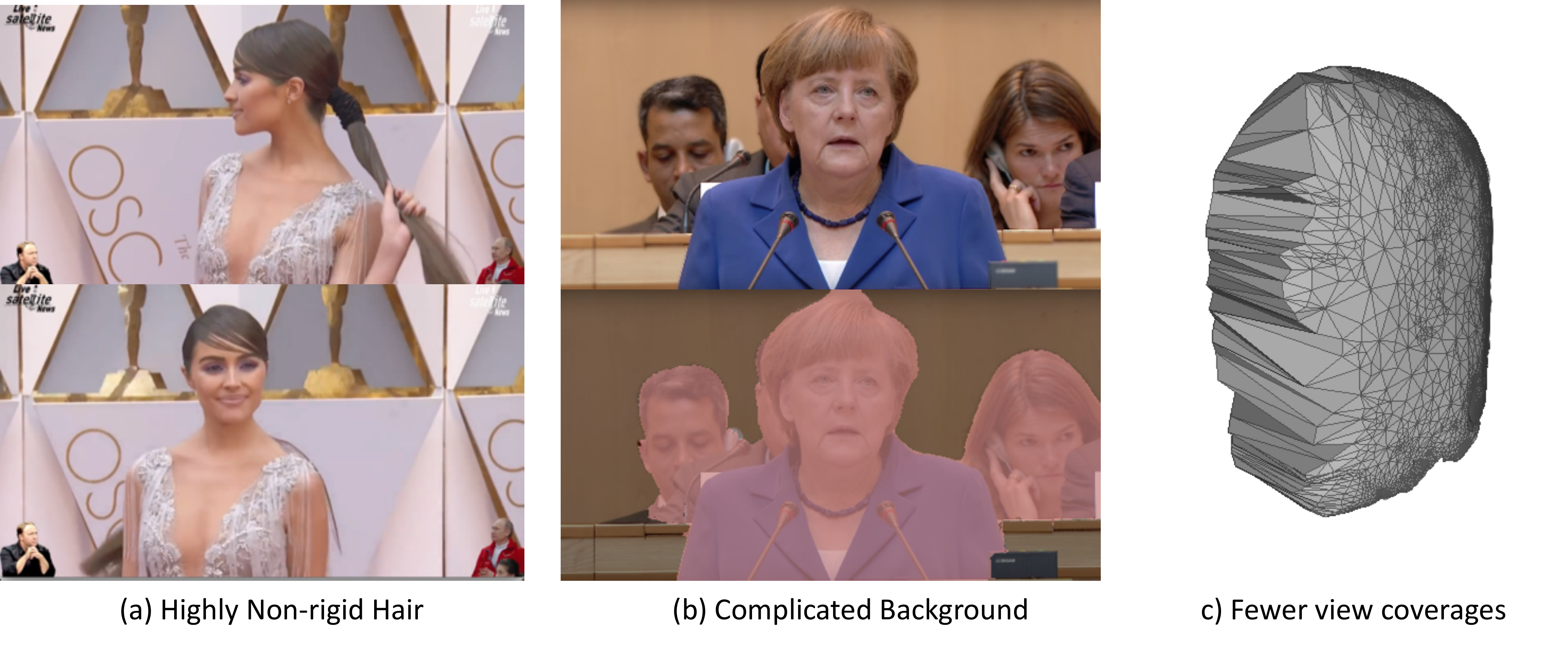}
  \caption{Limitations of our algorithm. In (a) we show example video frames of highly non-rigid hairstyle. In (b) we show an example video frame with a complicated background. In (c) we show the back of a deformed hair mesh towards a visual hull from a small view coverage input.}
  \label{fig:limitation}
\end{figure}

Our method cannot work on highly dynamic hairstyles due to the high non-rigidity of the hair volumes across the in-the-wild videos. See the example video frames of Olivia Culpo in Figure \ref{fig:limitation}(a). The human hand interaction and body occlusion make the segmentation difficult. Explicitly modeling long hair dynamics is beyond the scope of the paper, and we assume small dynamics from the input videos. We also could not reconstruct very curly hairstyles and complicated hairstyles such as braids.

Our method also fails on videos where the background is too complicated as shown in Figure \ref{fig:limitation}(b). The other people or crowds in the background make it hard to estimate the head silhouette of the person and will lead to incorrect correspondences when running structure-from-motion.

For the low-confidence view corrections, we require an input video covering a view range of at least $90$ degrees. Fewer views will cause the visual hull to be extremely distorted as shown in Figure \ref{fig:limitation}(c), where our deformable registration will fail with a large fitting error. Note that as the view coverage decreases, this problem will be reduced to a single-view reconstruction problem.

% Also, similarly to \cite{chai2016autohair,hu2015single}, our method won't perform well if the input hairstyle does not have a close match to any hairstyle in the database.

\subsection{Applications}
\begin{figure*}
  \includegraphics[width=\textwidth]{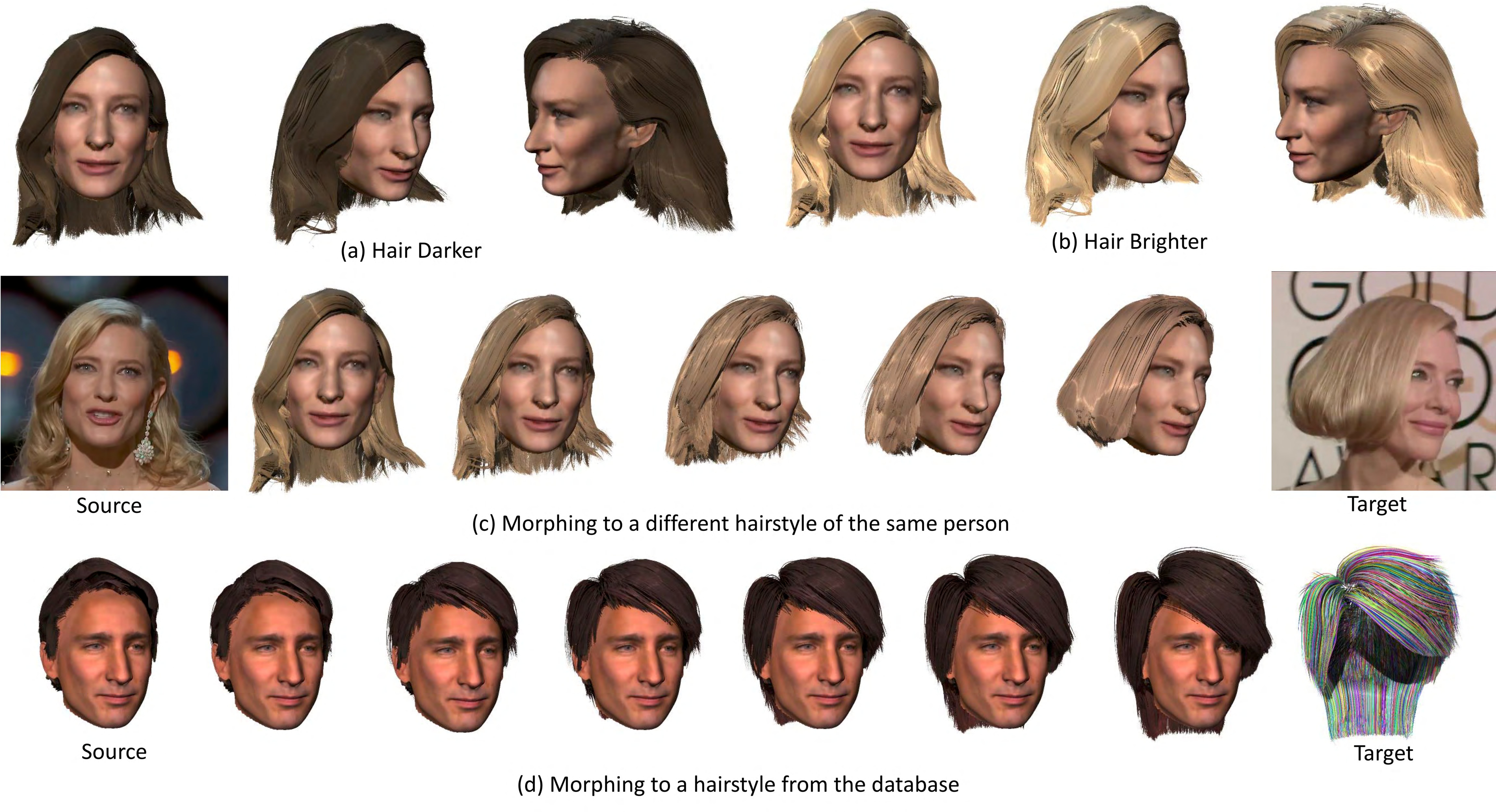}
  \caption{Hairstyle change examples. We show a darker and lighter version of Cate Blanchett's hairstyle in (a)(b). (c) shows the hair morphing intermediate results from two different hairstyles of the same person. (d) shows the hair morphing from the person's reconstructed hairstyle to a given hairstyle from the dataset.}
  \label{fig:morphing}
\end{figure*}

Our reconstructed models can be now used for a variety of applications as shown in Figure \ref{fig:morphing}(a)(b); we can also change the overall color of the hairstyle, making it darker or lighter, or morphing it to another hairstyle.

Since the hair roots of all our hair models are transfered from the generic shape used to compute the head model, we can assume that hair root points and hair strands are in correspondence for the same person. We resampled all the hair strands with the same number of vertices (50 in our implementations), which can be used for applications such as personalized strand-level hairstyle morphing.

In Figure \ref{fig:morphing}(c), we show the hair morphing result of Cate Blanchett in two hairstyles from two different videos. The intermediate results are created by a one-to-one strand interpolation of the source and target hair strands. We can also morph the reconstructed hairstyle to a given hairstyle from the dataset as shown in Figure \ref{fig:morphing}(d). The given hairstyle was re-grown from its 3D orientation field using the same set of scalp points to create correspondences to the original hairstyle. We trimmed the hair strands that intersect with the face during interpolation.

\section{Discussion and Future Work}
We have described a method that takes as input a video of a person's head in the wild and outputs a detailed 3D hair strand model combined with a reconstructed 3D head to produce a full head model. This method is fully automatic and shows that a head model with higher fidelity can be recovered by combining information from video frames. Our method is not restricted to specific views and head poses, making the full head reconstruction from in-the-wild videos possible. We showed our results on several celebrities as well as mobile selfie videos and compared our work to the most recent state of the art. 

However, there are still a number of limitations and possible extensions to explore.
One direction is to refine the rough hair mesh estimation for non-rigid hairstyles. Currently in our input, the person's head moves gently and our rough hair mesh is generated from the visual hull, which is only an approximation of the real hair volume, since the hair is non-rigid. We might explore using the ARkit of a smart phone to provide extra depth information to align the hair volume densely across frames. Also, we currently rely on the rigid camera poses estimated using structure from motion with SIFT features to connect all the views. In the future, we want to use facial features and hair specific features to increase the robustness of the frame alignment.

In our paper, we aim at a more challenging problem which is reconstructing 3D hair models from in-the-wild data, so some results are still not highly detailed, and we cannot handle complicated hairstyles such as braids and highly curled hairs.

We created face textures from video frames, which caused artifacts due to hair occlusions, decoration occlusions and low resolution of the faces. In the future, we can use generate photo-realistic textures as in \cite{saito2016photorealistic}.
Finally, a future extension to incorporate a facial blend shape model or estimate per frame facial dynamics as in \cite{suwajanakorn2014total} to create a fully animatable model would be desirable. 

\begin{acks}
This work was partially supported by the UW Reality Lab, Facebook, Google, Huawei, and NSF/Intel Visual and Experimental Computing Award \#1538618.
\end{acks}
\bibliographystyle{ACM-Reference-Format}
\bibliography{paper} 


\begin{thebibliography}{56}

%%% the Web address.

\ifx \showCODEN    \undefined \def \showCODEN     #1{\unskip}     \fi
\ifx \showDOI      \undefined \def \showDOI       #1{#1}\fi
\ifx \showISBNx    \undefined \def \showISBNx     #1{\unskip}     \fi
\ifx \showISBNxiii \undefined \def \showISBNxiii  #1{\unskip}     \fi
\ifx \showISSN     \undefined \def \showISSN      #1{\unskip}     \fi
\ifx \showLCCN     \undefined \def \showLCCN      #1{\unskip}     \fi
\ifx \shownote     \undefined \def \shownote      #1{#1}          \fi
\ifx \showarticletitle \undefined \def \showarticletitle #1{#1}   \fi
\ifx \showURL      \undefined \def \showURL       {\relax}        \fi
% invisible to TeX
\providecommand\bibfield[2]{#2}
\providecommand\bibinfo[2]{#2}
\providecommand\natexlab[1]{#1}
\providecommand\showeprint[2][]{arXiv:#2}

\bibitem[\protect\citeauthoryear{Alexander, Fyffe, Busch, Yu, Ichikari, Jones,
  Debevec, Jimenez, Danvoye, Antionazzi, et~al\mbox{.}}{Alexander
  et~al\mbox{.}}{2013}]%
        {alexander2013digital}
\bibfield{author}{\bibinfo{person}{Oleg Alexander}, \bibinfo{person}{Graham
  Fyffe}, \bibinfo{person}{Jay Busch}, \bibinfo{person}{Xueming Yu},
  \bibinfo{person}{Ryosuke Ichikari}, \bibinfo{person}{Andrew Jones},
  \bibinfo{person}{Paul Debevec}, \bibinfo{person}{Jorge Jimenez},
  \bibinfo{person}{Etienne Danvoye}, \bibinfo{person}{Bernardo Antionazzi},
  {et~al\mbox{.}}} \bibinfo{year}{2013}\natexlab{}.
\newblock \showarticletitle{Digital Ira: creating a real-time photoreal digital
  actor}. In \bibinfo{booktitle}{{\em ACM SIGGRAPH 2013 Posters}}. ACM,
  \bibinfo{pages}{1}.
\newblock


\bibitem[\protect\citeauthoryear{Allen, Curless, and Popovi{\'c}}{Allen
  et~al\mbox{.}}{2003}]%
        {allen}
\bibfield{author}{\bibinfo{person}{B. Allen}, \bibinfo{person}{B. Curless},
  {and} \bibinfo{person}{Z. Popovi{\'c}}.} \bibinfo{year}{2003}\natexlab{}.
\newblock \showarticletitle{The space of human body shapes: reconstruction and
  parameterization from range scans}. In \bibinfo{booktitle}{{\em ACM
  Transactions on Graphics (TOG)}}, Vol.~\bibinfo{volume}{22}. ACM,
  \bibinfo{pages}{587--594}.
\newblock


\bibitem[\protect\citeauthoryear{Beeler, Bickel, Beardsley, Sumner, and
  Gross}{Beeler et~al\mbox{.}}{2010}]%
        {beeler2010high}
\bibfield{author}{\bibinfo{person}{Thabo Beeler}, \bibinfo{person}{Bernd
  Bickel}, \bibinfo{person}{Paul Beardsley}, \bibinfo{person}{Bob Sumner},
  {and} \bibinfo{person}{Markus Gross}.} \bibinfo{year}{2010}\natexlab{}.
\newblock \showarticletitle{High-quality single-shot capture of facial
  geometry}.
\newblock \bibinfo{journal}{{\em ACM Transactions on Graphics (TOG)\/}}
  \bibinfo{volume}{29}, \bibinfo{number}{4} (\bibinfo{year}{2010}),
  \bibinfo{pages}{40}.
\newblock


\bibitem[\protect\citeauthoryear{Blanz and Vetter}{Blanz and Vetter}{1999}]%
        {blanz1999morphable}
\bibfield{author}{\bibinfo{person}{Volker Blanz} {and} \bibinfo{person}{Thomas
  Vetter}.} \bibinfo{year}{1999}\natexlab{}.
\newblock \showarticletitle{A morphable model for the synthesis of 3D faces}.
  In \bibinfo{booktitle}{{\em Proceedings of the 26th annual conference on
  Computer graphics and interactive techniques}}. ACM Press/Addison-Wesley
  Publishing Co., \bibinfo{pages}{187--194}.
\newblock


\bibitem[\protect\citeauthoryear{Cao, Weng, Zhou, Tong, and Zhou}{Cao
  et~al\mbox{.}}{2013}]%
        {cao2013facewarehouse}
\bibfield{author}{\bibinfo{person}{Chen Cao}, \bibinfo{person}{Yanlin Weng},
  \bibinfo{person}{Shun Zhou}, \bibinfo{person}{Yiying Tong}, {and}
  \bibinfo{person}{Kun Zhou}.} \bibinfo{year}{2013}\natexlab{}.
\newblock \showarticletitle{Facewarehouse: a 3d facial expression database for
  visual computing}.
\newblock  (\bibinfo{year}{2013}).
\newblock


\bibitem[\protect\citeauthoryear{Cao, Wu, Weng, Shao, and Zhou}{Cao
  et~al\mbox{.}}{2016}]%
        {cao2016real}
\bibfield{author}{\bibinfo{person}{Chen Cao}, \bibinfo{person}{Hongzhi Wu},
  \bibinfo{person}{Yanlin Weng}, \bibinfo{person}{Tianjia Shao}, {and}
  \bibinfo{person}{Kun Zhou}.} \bibinfo{year}{2016}\natexlab{}.
\newblock \showarticletitle{Real-time facial animation with image-based dynamic
  avatars}.
\newblock \bibinfo{journal}{{\em ACM Transactions on Graphics (TOG)\/}}
  \bibinfo{volume}{35}, \bibinfo{number}{4} (\bibinfo{year}{2016}),
  \bibinfo{pages}{126}.
\newblock


\bibitem[\protect\citeauthoryear{Chai, Luo, Sunkavalli, Carr, Hadap, and
  Zhou}{Chai et~al\mbox{.}}{2015}]%
        {chai2015high}
\bibfield{author}{\bibinfo{person}{Menglei Chai}, \bibinfo{person}{Linjie Luo},
  \bibinfo{person}{Kalyan Sunkavalli}, \bibinfo{person}{Nathan Carr},
  \bibinfo{person}{Sunil Hadap}, {and} \bibinfo{person}{Kun Zhou}.}
  \bibinfo{year}{2015}\natexlab{}.
\newblock \showarticletitle{High-quality hair modeling from a single portrait
  photo}.
\newblock \bibinfo{journal}{{\em ACM Transactions on Graphics (TOG)\/}}
  \bibinfo{volume}{34}, \bibinfo{number}{6} (\bibinfo{year}{2015}),
  \bibinfo{pages}{204}.
\newblock


\bibitem[\protect\citeauthoryear{Chai, Shao, Wu, Weng, and Zhou}{Chai
  et~al\mbox{.}}{2016}]%
        {chai2016autohair}
\bibfield{author}{\bibinfo{person}{Menglei Chai}, \bibinfo{person}{Tianjia
  Shao}, \bibinfo{person}{Hongzhi Wu}, \bibinfo{person}{Yanlin Weng}, {and}
  \bibinfo{person}{Kun Zhou}.} \bibinfo{year}{2016}\natexlab{}.
\newblock \showarticletitle{Autohair: Fully automatic hair modeling from a
  single image}.
\newblock \bibinfo{journal}{{\em ACM Transactions on Graphics (TOG)\/}}
  \bibinfo{volume}{35}, \bibinfo{number}{4} (\bibinfo{year}{2016}),
  \bibinfo{pages}{116}.
\newblock


\bibitem[\protect\citeauthoryear{Chai, Wang, Weng, Jin, and Zhou}{Chai
  et~al\mbox{.}}{2013}]%
        {chai2013dynamic}
\bibfield{author}{\bibinfo{person}{Menglei Chai}, \bibinfo{person}{Lvdi Wang},
  \bibinfo{person}{Yanlin Weng}, \bibinfo{person}{Xiaogang Jin}, {and}
  \bibinfo{person}{Kun Zhou}.} \bibinfo{year}{2013}\natexlab{}.
\newblock \showarticletitle{Dynamic hair manipulation in images and videos}.
\newblock \bibinfo{journal}{{\em ACM Transactions on Graphics (TOG)\/}}
  \bibinfo{volume}{32}, \bibinfo{number}{4} (\bibinfo{year}{2013}),
  \bibinfo{pages}{75}.
\newblock


\bibitem[\protect\citeauthoryear{Chai, Wang, Weng, Yu, Guo, and Zhou}{Chai
  et~al\mbox{.}}{2012}]%
        {chai2012single}
\bibfield{author}{\bibinfo{person}{Menglei Chai}, \bibinfo{person}{Lvdi Wang},
  \bibinfo{person}{Yanlin Weng}, \bibinfo{person}{Yizhou Yu},
  \bibinfo{person}{Baining Guo}, {and} \bibinfo{person}{Kun Zhou}.}
  \bibinfo{year}{2012}\natexlab{}.
\newblock \showarticletitle{Single-view hair modeling for portrait
  manipulation}.
\newblock \bibinfo{journal}{{\em ACM Transactions on Graphics (TOG)\/}}
  \bibinfo{volume}{31}, \bibinfo{number}{4} (\bibinfo{year}{2012}),
  \bibinfo{pages}{116}.
\newblock


\bibitem[\protect\citeauthoryear{Chen, Papandreou, Kokkinos, Murphy, and
  Yuille}{Chen et~al\mbox{.}}{2016}]%
        {chen2016deeplab}
\bibfield{author}{\bibinfo{person}{Liang-Chieh Chen}, \bibinfo{person}{George
  Papandreou}, \bibinfo{person}{Iasonas Kokkinos}, \bibinfo{person}{Kevin
  Murphy}, {and} \bibinfo{person}{Alan~L Yuille}.}
  \bibinfo{year}{2016}\natexlab{}.
\newblock \showarticletitle{Deeplab: Semantic image segmentation with deep
  convolutional nets, atrous convolution, and fully connected crfs}.
\newblock \bibinfo{journal}{{\em arXiv preprint arXiv:1606.00915\/}}
  (\bibinfo{year}{2016}).
\newblock


\bibitem[\protect\citeauthoryear{Debevec}{Debevec}{2012}]%
        {debevec2012light}
\bibfield{author}{\bibinfo{person}{Paul Debevec}.}
  \bibinfo{year}{2012}\natexlab{}.
\newblock \showarticletitle{The light stages and their applications to
  photoreal digital actors}.
\newblock \bibinfo{journal}{{\em SIGGRAPH Asia\/}} (\bibinfo{year}{2012}).
\newblock


\bibitem[\protect\citeauthoryear{Fu, Wei, Tai, and Quan}{Fu
  et~al\mbox{.}}{2007}]%
        {fu2007sketching}
\bibfield{author}{\bibinfo{person}{Hongbo Fu}, \bibinfo{person}{Yichen Wei},
  \bibinfo{person}{Chiew-Lan Tai}, {and} \bibinfo{person}{Long Quan}.}
  \bibinfo{year}{2007}\natexlab{}.
\newblock \showarticletitle{Sketching hairstyles}. In \bibinfo{booktitle}{{\em
  Proceedings of the 4th Eurographics workshop on Sketch-based interfaces and
  modeling}}. ACM, \bibinfo{pages}{31--36}.
\newblock


\bibitem[\protect\citeauthoryear{Goesele, Snavely, Curless, Hoppe, and
  Seitz}{Goesele et~al\mbox{.}}{2007}]%
        {goesele2007multi}
\bibfield{author}{\bibinfo{person}{Michael Goesele}, \bibinfo{person}{Noah
  Snavely}, \bibinfo{person}{Brian Curless}, \bibinfo{person}{Hugues Hoppe},
  {and} \bibinfo{person}{Steven~M Seitz}.} \bibinfo{year}{2007}\natexlab{}.
\newblock \showarticletitle{Multi-view stereo for community photo collections}.
  In \bibinfo{booktitle}{{\em Computer Vision, 2007. ICCV 2007. IEEE 11th
  International Conference on}}. IEEE, \bibinfo{pages}{1--8}.
\newblock


\bibitem[\protect\citeauthoryear{Hairbobo}{Hairbobo}{2017}]%
        {Hairbobo}
\bibfield{author}{\bibinfo{person}{Hairbobo}.} \bibinfo{year}{2017}\natexlab{}.
\newblock \bibinfo{title}{Hairbobo}.
\newblock \bibinfo{howpublished}{\url{http://www.hairbobo.com/faxingtupian}}.
  (\bibinfo{date}{Sept.} \bibinfo{year}{2017}).
\newblock


\bibitem[\protect\citeauthoryear{Hu, Ma, Luo, and Li}{Hu
  et~al\mbox{.}}{2014a}]%
        {hu2014robust}
\bibfield{author}{\bibinfo{person}{Liwen Hu}, \bibinfo{person}{Chongyang Ma},
  \bibinfo{person}{Linjie Luo}, {and} \bibinfo{person}{Hao Li}.}
  \bibinfo{year}{2014}\natexlab{a}.
\newblock \showarticletitle{Robust hair capture using simulated examples}.
\newblock \bibinfo{journal}{{\em ACM Transactions on Graphics (TOG)\/}}
  \bibinfo{volume}{33}, \bibinfo{number}{4} (\bibinfo{year}{2014}),
  \bibinfo{pages}{126}.
\newblock


\bibitem[\protect\citeauthoryear{Hu, Ma, Luo, and Li}{Hu et~al\mbox{.}}{2015}]%
        {hu2015single}
\bibfield{author}{\bibinfo{person}{Liwen Hu}, \bibinfo{person}{Chongyang Ma},
  \bibinfo{person}{Linjie Luo}, {and} \bibinfo{person}{Hao Li}.}
  \bibinfo{year}{2015}\natexlab{}.
\newblock \showarticletitle{Single-view hair modeling using a hairstyle
  database}.
\newblock \bibinfo{journal}{{\em ACM Transactions on Graphics (TOG)\/}}
  \bibinfo{volume}{34}, \bibinfo{number}{4} (\bibinfo{year}{2015}),
  \bibinfo{pages}{125}.
\newblock


\bibitem[\protect\citeauthoryear{Hu, Ma, Luo, Wei, and Li}{Hu
  et~al\mbox{.}}{2014b}]%
        {hu2014capturing}
\bibfield{author}{\bibinfo{person}{Liwen Hu}, \bibinfo{person}{Chongyang Ma},
  \bibinfo{person}{Linjie Luo}, \bibinfo{person}{Li-Yi Wei}, {and}
  \bibinfo{person}{Hao Li}.} \bibinfo{year}{2014}\natexlab{b}.
\newblock \showarticletitle{Capturing braided hairstyles}.
\newblock \bibinfo{journal}{{\em ACM Transactions on Graphics (TOG)\/}}
  \bibinfo{volume}{33}, \bibinfo{number}{6} (\bibinfo{year}{2014}),
  \bibinfo{pages}{225}.
\newblock


\bibitem[\protect\citeauthoryear{Hu, Saito, Wei, Nagano, Seo, Fursund, Sadeghi,
  Sun, Chen, and Li}{Hu et~al\mbox{.}}{2017}]%
        {Hu:2017:ADS:3130800.31310887}
\bibfield{author}{\bibinfo{person}{Liwen Hu}, \bibinfo{person}{Shunsuke Saito},
  \bibinfo{person}{Lingyu Wei}, \bibinfo{person}{Koki Nagano},
  \bibinfo{person}{Jaewoo Seo}, \bibinfo{person}{Jens Fursund},
  \bibinfo{person}{Iman Sadeghi}, \bibinfo{person}{Carrie Sun},
  \bibinfo{person}{Yen-Chun Chen}, {and} \bibinfo{person}{Hao Li}.}
  \bibinfo{year}{2017}\natexlab{}.
\newblock \showarticletitle{Avatar Digitization from a Single Image for
  Real-time Rendering}.
\newblock \bibinfo{journal}{{\em ACM Trans. Graph.\/}} \bibinfo{volume}{36},
  \bibinfo{number}{6}, Article \bibinfo{articleno}{195} (\bibinfo{date}{Nov.}
  \bibinfo{year}{2017}), \bibinfo{numpages}{14}~pages.
\newblock
\showISSN{0730-0301}
\showDOI{%
\url{https://doi.org/10.1145/3130800.31310887}}


\bibitem[\protect\citeauthoryear{Ichim, Bouaziz, and Pauly}{Ichim
  et~al\mbox{.}}{2015}]%
        {ichim2015dynamic}
\bibfield{author}{\bibinfo{person}{Alexandru~Eugen Ichim},
  \bibinfo{person}{Sofien Bouaziz}, {and} \bibinfo{person}{Mark Pauly}.}
  \bibinfo{year}{2015}\natexlab{}.
\newblock \showarticletitle{Dynamic 3D avatar creation from hand-held video
  input}.
\newblock \bibinfo{journal}{{\em ACM Transactions on Graphics (TOG)\/}}
  \bibinfo{volume}{34}, \bibinfo{number}{4} (\bibinfo{year}{2015}),
  \bibinfo{pages}{45}.
\newblock


\bibitem[\protect\citeauthoryear{Jakob, Moon, and Marschner}{Jakob
  et~al\mbox{.}}{2009}]%
        {jakob2009capturing}
\bibfield{author}{\bibinfo{person}{Wenzel Jakob}, \bibinfo{person}{Jonathan~T
  Moon}, {and} \bibinfo{person}{Steve Marschner}.}
  \bibinfo{year}{2009}\natexlab{}.
\newblock \showarticletitle{Capturing hair assemblies fiber by fiber}. In
  \bibinfo{booktitle}{{\em ACM Transactions on Graphics (TOG)}},
  Vol.~\bibinfo{volume}{28}. ACM, \bibinfo{pages}{164}.
\newblock


\bibitem[\protect\citeauthoryear{Jia, Shelhamer, Donahue, Karayev, Long,
  Girshick, Guadarrama, and Darrell}{Jia et~al\mbox{.}}{2014}]%
        {jia2014caffe}
\bibfield{author}{\bibinfo{person}{Yangqing Jia}, \bibinfo{person}{Evan
  Shelhamer}, \bibinfo{person}{Jeff Donahue}, \bibinfo{person}{Sergey Karayev},
  \bibinfo{person}{Jonathan Long}, \bibinfo{person}{Ross Girshick},
  \bibinfo{person}{Sergio Guadarrama}, {and} \bibinfo{person}{Trevor Darrell}.}
  \bibinfo{year}{2014}\natexlab{}.
\newblock \showarticletitle{Caffe: Convolutional architecture for fast feature
  embedding}. In \bibinfo{booktitle}{{\em Proceedings of the 22nd ACM
  international conference on Multimedia}}. ACM, \bibinfo{pages}{675--678}.
\newblock


\bibitem[\protect\citeauthoryear{Kazhdan and Hoppe}{Kazhdan and Hoppe}{2013}]%
        {kazhdan2013screened}
\bibfield{author}{\bibinfo{person}{Michael Kazhdan} {and}
  \bibinfo{person}{Hugues Hoppe}.} \bibinfo{year}{2013}\natexlab{}.
\newblock \showarticletitle{Screened poisson surface reconstruction}.
\newblock \bibinfo{journal}{{\em ACM Transactions on Graphics (TOG)\/}}
  \bibinfo{volume}{32}, \bibinfo{number}{3} (\bibinfo{year}{2013}),
  \bibinfo{pages}{29}.
\newblock


\bibitem[\protect\citeauthoryear{Kemelmacher-Shlizerman}{Kemelmacher-Shlizerman}{2016}]%
        {kemelmacher2016transfiguring}
\bibfield{author}{\bibinfo{person}{Ira Kemelmacher-Shlizerman}.}
  \bibinfo{year}{2016}\natexlab{}.
\newblock \showarticletitle{Transfiguring portraits}.
\newblock \bibinfo{journal}{{\em ACM Transactions on Graphics (TOG)\/}}
  \bibinfo{volume}{35}, \bibinfo{number}{4} (\bibinfo{year}{2016}),
  \bibinfo{pages}{94}.
\newblock


\bibitem[\protect\citeauthoryear{Kemelmacher-Shlizerman and
  Basri}{Kemelmacher-Shlizerman and Basri}{2011}]%
        {kemelmacher20113d}
\bibfield{author}{\bibinfo{person}{Ira Kemelmacher-Shlizerman} {and}
  \bibinfo{person}{Ronen Basri}.} \bibinfo{year}{2011}\natexlab{}.
\newblock \showarticletitle{3d face reconstruction from a single image using a
  single reference face shape}.
\newblock \bibinfo{journal}{{\em Pattern Analysis and Machine Intelligence,
  IEEE Transactions on\/}} \bibinfo{volume}{33}, \bibinfo{number}{2}
  (\bibinfo{year}{2011}), \bibinfo{pages}{394--405}.
\newblock


\bibitem[\protect\citeauthoryear{Kemelmacher-Shlizerman and
  Seitz}{Kemelmacher-Shlizerman and Seitz}{2011}]%
        {kemelmacher2011face}
\bibfield{author}{\bibinfo{person}{Ira Kemelmacher-Shlizerman} {and}
  \bibinfo{person}{Steven~M Seitz}.} \bibinfo{year}{2011}\natexlab{}.
\newblock \showarticletitle{Face reconstruction in the wild}. In
  \bibinfo{booktitle}{{\em Computer Vision (ICCV), 2011 IEEE International
  Conference on}}. IEEE, \bibinfo{pages}{1746--1753}.
\newblock


\bibitem[\protect\citeauthoryear{Laurentini}{Laurentini}{1994}]%
        {laurentini1994visual}
\bibfield{author}{\bibinfo{person}{Aldo Laurentini}.}
  \bibinfo{year}{1994}\natexlab{}.
\newblock \showarticletitle{The visual hull concept for silhouette-based image
  understanding}.
\newblock \bibinfo{journal}{{\em IEEE Transactions on pattern analysis and
  machine intelligence\/}} \bibinfo{volume}{16}, \bibinfo{number}{2}
  (\bibinfo{year}{1994}), \bibinfo{pages}{150--162}.
\newblock


\bibitem[\protect\citeauthoryear{Liang, Kemelmacher-Shlizerman, and
  Shapiro}{Liang et~al\mbox{.}}{2014}]%
        {liang20143d}
\bibfield{author}{\bibinfo{person}{Shu Liang}, \bibinfo{person}{Ira
  Kemelmacher-Shlizerman}, {and} \bibinfo{person}{Linda~G Shapiro}.}
  \bibinfo{year}{2014}\natexlab{}.
\newblock \showarticletitle{3d face hallucination from a single depth frame}.
  In \bibinfo{booktitle}{{\em 3D Vision (3DV), 2014 2nd international
  conference on}}, Vol.~\bibinfo{volume}{1}. IEEE, \bibinfo{pages}{31--38}.
\newblock


\bibitem[\protect\citeauthoryear{Liang, Shapiro, and
  Kemelmacher-Shlizerman}{Liang et~al\mbox{.}}{2016}]%
        {liang2016head}
\bibfield{author}{\bibinfo{person}{Shu Liang}, \bibinfo{person}{Linda~G
  Shapiro}, {and} \bibinfo{person}{Ira Kemelmacher-Shlizerman}.}
  \bibinfo{year}{2016}\natexlab{}.
\newblock \showarticletitle{Head reconstruction from internet photos}. In
  \bibinfo{booktitle}{{\em European Conference on Computer Vision}}. Springer,
  \bibinfo{pages}{360--374}.
\newblock


\bibitem[\protect\citeauthoryear{Liu, Shi, Liang, and Yang}{Liu
  et~al\mbox{.}}{2017}]%
        {liu2017face}
\bibfield{author}{\bibinfo{person}{Sifei Liu}, \bibinfo{person}{Jianping Shi},
  \bibinfo{person}{Ji Liang}, {and} \bibinfo{person}{Ming-Hsuan Yang}.}
  \bibinfo{year}{2017}\natexlab{}.
\newblock \showarticletitle{Face Parsing via Recurrent Propagation}.
\newblock \bibinfo{journal}{{\em arXiv preprint arXiv:1708.01936\/}}
  (\bibinfo{year}{2017}).
\newblock


\bibitem[\protect\citeauthoryear{Liu, Yang, Huang, and Yang}{Liu
  et~al\mbox{.}}{2015}]%
        {liu2015multi}
\bibfield{author}{\bibinfo{person}{Sifei Liu}, \bibinfo{person}{Jimei Yang},
  \bibinfo{person}{Chang Huang}, {and} \bibinfo{person}{Ming-Hsuan Yang}.}
  \bibinfo{year}{2015}\natexlab{}.
\newblock \showarticletitle{Multi-objective convolutional learning for face
  labeling}. In \bibinfo{booktitle}{{\em Proceedings of the IEEE Conference on
  Computer Vision and Pattern Recognition}}. \bibinfo{pages}{3451--3459}.
\newblock


\bibitem[\protect\citeauthoryear{Long, Shelhamer, and Darrell}{Long
  et~al\mbox{.}}{2015}]%
        {long2015fully}
\bibfield{author}{\bibinfo{person}{Jonathan Long}, \bibinfo{person}{Evan
  Shelhamer}, {and} \bibinfo{person}{Trevor Darrell}.}
  \bibinfo{year}{2015}\natexlab{}.
\newblock \showarticletitle{Fully convolutional networks for semantic
  segmentation}. In \bibinfo{booktitle}{{\em Proceedings of the IEEE Conference
  on Computer Vision and Pattern Recognition}}. \bibinfo{pages}{3431--3440}.
\newblock


\bibitem[\protect\citeauthoryear{Luo, Li, and Rusinkiewicz}{Luo
  et~al\mbox{.}}{2013}]%
        {luo2013structure}
\bibfield{author}{\bibinfo{person}{Linjie Luo}, \bibinfo{person}{Hao Li}, {and}
  \bibinfo{person}{Szymon Rusinkiewicz}.} \bibinfo{year}{2013}\natexlab{}.
\newblock \showarticletitle{Structure-aware hair capture}.
\newblock \bibinfo{journal}{{\em ACM Transactions on Graphics (TOG)\/}}
  \bibinfo{volume}{32}, \bibinfo{number}{4} (\bibinfo{year}{2013}),
  \bibinfo{pages}{76}.
\newblock


\bibitem[\protect\citeauthoryear{Luo, Wang, and Tang}{Luo
  et~al\mbox{.}}{2012}]%
        {luo2012hierarchical}
\bibfield{author}{\bibinfo{person}{Ping Luo}, \bibinfo{person}{Xiaogang Wang},
  {and} \bibinfo{person}{Xiaoou Tang}.} \bibinfo{year}{2012}\natexlab{}.
\newblock \showarticletitle{Hierarchical face parsing via deep learning}. In
  \bibinfo{booktitle}{{\em Computer Vision and Pattern Recognition (CVPR), 2012
  IEEE Conference on}}. IEEE, \bibinfo{pages}{2480--2487}.
\newblock


\bibitem[\protect\citeauthoryear{Maninchedda, H{\"a}ne, Jacquet, Delaunoy, and
  Pollefeys}{Maninchedda et~al\mbox{.}}{2016}]%
        {maninchedda2016semantic}
\bibfield{author}{\bibinfo{person}{Fabio Maninchedda},
  \bibinfo{person}{Christian H{\"a}ne}, \bibinfo{person}{Bastien Jacquet},
  \bibinfo{person}{Ama{\"e}l Delaunoy}, {and} \bibinfo{person}{Marc
  Pollefeys}.} \bibinfo{year}{2016}\natexlab{}.
\newblock \showarticletitle{Semantic 3D Reconstruction of Heads}. In
  \bibinfo{booktitle}{{\em European Conference on Computer Vision}}. Springer,
  \bibinfo{pages}{667--683}.
\newblock


\bibitem[\protect\citeauthoryear{Newcombe, Fox, and Seitz}{Newcombe
  et~al\mbox{.}}{2015}]%
        {newcombe2015dynamicfusion}
\bibfield{author}{\bibinfo{person}{Richard~A Newcombe}, \bibinfo{person}{Dieter
  Fox}, {and} \bibinfo{person}{Steven~M Seitz}.}
  \bibinfo{year}{2015}\natexlab{}.
\newblock \showarticletitle{DynamicFusion: Reconstruction and tracking of
  non-rigid scenes in real-time}. In \bibinfo{booktitle}{{\em Proceedings of
  the IEEE Conference on Computer Vision and Pattern Recognition}}.
  \bibinfo{pages}{343--352}.
\newblock


\bibitem[\protect\citeauthoryear{Paris, Brice{\~n}o, and Sillion}{Paris
  et~al\mbox{.}}{2004}]%
        {paris2004capture}
\bibfield{author}{\bibinfo{person}{Sylvain Paris}, \bibinfo{person}{Hector~M
  Brice{\~n}o}, {and} \bibinfo{person}{Fran{\c{c}}ois~X Sillion}.}
  \bibinfo{year}{2004}\natexlab{}.
\newblock \showarticletitle{Capture of hair geometry from multiple images}. In
  \bibinfo{booktitle}{{\em ACM Transactions on Graphics (TOG)}},
  Vol.~\bibinfo{volume}{23}. ACM, \bibinfo{pages}{712--719}.
\newblock


\bibitem[\protect\citeauthoryear{Paris, Chang, Kozhushnyan, Jarosz, Matusik,
  Zwicker, and Durand}{Paris et~al\mbox{.}}{2008}]%
        {paris2008hair}
\bibfield{author}{\bibinfo{person}{Sylvain Paris}, \bibinfo{person}{Will
  Chang}, \bibinfo{person}{Oleg~I Kozhushnyan}, \bibinfo{person}{Wojciech
  Jarosz}, \bibinfo{person}{Wojciech Matusik}, \bibinfo{person}{Matthias
  Zwicker}, {and} \bibinfo{person}{Fr{\'e}do Durand}.}
  \bibinfo{year}{2008}\natexlab{}.
\newblock \showarticletitle{Hair photobooth: geometric and photometric
  acquisition of real hairstyles}. In \bibinfo{booktitle}{{\em ACM Transactions
  on Graphics (TOG)}}, Vol.~\bibinfo{volume}{27}. ACM, \bibinfo{pages}{30}.
\newblock


\bibitem[\protect\citeauthoryear{Richardson, Sela, and Kimmel}{Richardson
  et~al\mbox{.}}{2016}]%
        {richardson20163d}
\bibfield{author}{\bibinfo{person}{Elad Richardson}, \bibinfo{person}{Matan
  Sela}, {and} \bibinfo{person}{Ron Kimmel}.} \bibinfo{year}{2016}\natexlab{}.
\newblock \showarticletitle{3D face reconstruction by learning from synthetic
  data}. In \bibinfo{booktitle}{{\em 3D Vision (3DV), 2016 Fourth International
  Conference on}}. IEEE, \bibinfo{pages}{460--469}.
\newblock


\bibitem[\protect\citeauthoryear{Richardson, Sela, Or-El, and
  Kimmel}{Richardson et~al\mbox{.}}{2017}]%
        {richardson2017learning}
\bibfield{author}{\bibinfo{person}{Elad Richardson}, \bibinfo{person}{Matan
  Sela}, \bibinfo{person}{Roy Or-El}, {and} \bibinfo{person}{Ron Kimmel}.}
  \bibinfo{year}{2017}\natexlab{}.
\newblock \showarticletitle{Learning detailed face reconstruction from a single
  image}. In \bibinfo{booktitle}{{\em 2017 IEEE Conference on Computer Vision
  and Pattern Recognition (CVPR)}}. IEEE, \bibinfo{pages}{5553--5562}.
\newblock


\bibitem[\protect\citeauthoryear{Saito, Wei, Hu, Nagano, and Li}{Saito
  et~al\mbox{.}}{2016}]%
        {saito2016photorealistic}
\bibfield{author}{\bibinfo{person}{Shunsuke Saito}, \bibinfo{person}{Lingyu
  Wei}, \bibinfo{person}{Liwen Hu}, \bibinfo{person}{Koki Nagano}, {and}
  \bibinfo{person}{Hao Li}.} \bibinfo{year}{2016}\natexlab{}.
\newblock \showarticletitle{Photorealistic Facial Texture Inference Using Deep
  Neural Networks}.
\newblock \bibinfo{journal}{{\em arXiv preprint arXiv:1612.00523\/}}
  (\bibinfo{year}{2016}).
\newblock


\bibitem[\protect\citeauthoryear{Sorkine, Cohen-Or, Lipman, Alexa, R{\"o}ssl,
  and Seidel}{Sorkine et~al\mbox{.}}{2004}]%
        {sorkine2004laplacian}
\bibfield{author}{\bibinfo{person}{Olga Sorkine}, \bibinfo{person}{Daniel
  Cohen-Or}, \bibinfo{person}{Yaron Lipman}, \bibinfo{person}{Marc Alexa},
  \bibinfo{person}{Christian R{\"o}ssl}, {and} \bibinfo{person}{H-P Seidel}.}
  \bibinfo{year}{2004}\natexlab{}.
\newblock \showarticletitle{Laplacian surface editing}. In
  \bibinfo{booktitle}{{\em Proceedings of the 2004 Eurographics/ACM SIGGRAPH
  symposium on Geometry processing}}. ACM, \bibinfo{pages}{175--184}.
\newblock


\bibitem[\protect\citeauthoryear{Suwajanakorn, Kemelmacher-Shlizerman, and
  Seitz}{Suwajanakorn et~al\mbox{.}}{2014}]%
        {suwajanakorn2014total}
\bibfield{author}{\bibinfo{person}{Supasorn Suwajanakorn}, \bibinfo{person}{Ira
  Kemelmacher-Shlizerman}, {and} \bibinfo{person}{Steven~M Seitz}.}
  \bibinfo{year}{2014}\natexlab{}.
\newblock \showarticletitle{Total moving face reconstruction}.
\newblock In \bibinfo{booktitle}{{\em Computer Vision--ECCV 2014}}.
\newblock


\bibitem[\protect\citeauthoryear{Suwajanakorn, Seitz, and
  Kemelmacher-Shlizerman}{Suwajanakorn et~al\mbox{.}}{2015}]%
        {suwajanakorn2015makes}
\bibfield{author}{\bibinfo{person}{Supasorn Suwajanakorn},
  \bibinfo{person}{Steven~M Seitz}, {and} \bibinfo{person}{Ira
  Kemelmacher-Shlizerman}.} \bibinfo{year}{2015}\natexlab{}.
\newblock \showarticletitle{What Makes Tom Hanks Look Like Tom Hanks}. In
  \bibinfo{booktitle}{{\em Proceedings of the IEEE International Conference on
  Computer Vision}}. \bibinfo{pages}{3952--3960}.
\newblock


\bibitem[\protect\citeauthoryear{Thies, Zollhoefer, Niessner, Valgaerts,
  Stamminger, and Theobalt}{Thies et~al\mbox{.}}{2015}]%
        {thies2015real}
\bibfield{author}{\bibinfo{person}{Justus Thies}, \bibinfo{person}{Michael
  Zollhoefer}, \bibinfo{person}{Matthias Niessner}, \bibinfo{person}{Levi
  Valgaerts}, \bibinfo{person}{Marc Stamminger}, {and}
  \bibinfo{person}{Christian Theobalt}.} \bibinfo{year}{2015}\natexlab{}.
\newblock \showarticletitle{Real-time Expression Transfer for Facial
  Reenactment}.
\newblock \bibinfo{journal}{{\em ACM Transactions on Graphics (Proc. SIGGRAPH
  Asia)\/}} (\bibinfo{year}{2015}).
\newblock


\bibitem[\protect\citeauthoryear{Tran, Hassner, Masi, and Medioni}{Tran
  et~al\mbox{.}}{2017}]%
        {tran2017regressing}
\bibfield{author}{\bibinfo{person}{Anh~Tuan Tran}, \bibinfo{person}{Tal
  Hassner}, \bibinfo{person}{Iacopo Masi}, {and} \bibinfo{person}{G{\'e}rard
  Medioni}.} \bibinfo{year}{2017}\natexlab{}.
\newblock \showarticletitle{Regressing robust and discriminative 3D morphable
  models with a very deep neural network}. In \bibinfo{booktitle}{{\em 2017
  IEEE Conference on Computer Vision and Pattern Recognition (CVPR)}}. IEEE,
  \bibinfo{pages}{1493--1502}.
\newblock


\bibitem[\protect\citeauthoryear{Vanakittistien, Sudsang, and
  Chentanez}{Vanakittistien et~al\mbox{.}}{2016}]%
        {vanakittistien20163d}
\bibfield{author}{\bibinfo{person}{Nuttapon Vanakittistien},
  \bibinfo{person}{Attawith Sudsang}, {and} \bibinfo{person}{Nuttapong
  Chentanez}.} \bibinfo{year}{2016}\natexlab{}.
\newblock \showarticletitle{3D hair model from small set of images}. In
  \bibinfo{booktitle}{{\em Proceedings of the 9th International Conference on
  Motion in Games}}. ACM, \bibinfo{pages}{85--90}.
\newblock


\bibitem[\protect\citeauthoryear{Wang, Chai, Zhang, Chang, Zeng, and Shan}{Wang
  et~al\mbox{.}}{2011}]%
        {wang2011novel}
\bibfield{author}{\bibinfo{person}{Dan Wang}, \bibinfo{person}{Xiujuan Chai},
  \bibinfo{person}{Hongming Zhang}, \bibinfo{person}{Hong Chang},
  \bibinfo{person}{Wei Zeng}, {and} \bibinfo{person}{Shiguang Shan}.}
  \bibinfo{year}{2011}\natexlab{}.
\newblock \showarticletitle{A novel coarse-to-fine hair segmentation method}.
  In \bibinfo{booktitle}{{\em Automatic Face \& Gesture Recognition and
  Workshops (FG 2011), 2011 IEEE International Conference on}}. IEEE,
  \bibinfo{pages}{233--238}.
\newblock


\bibitem[\protect\citeauthoryear{Ward, Bertails, Kim, Marschner, Cani, and
  Lin}{Ward et~al\mbox{.}}{2007}]%
        {ward2007survey}
\bibfield{author}{\bibinfo{person}{Kelly Ward}, \bibinfo{person}{Florence
  Bertails}, \bibinfo{person}{Tae-Yong Kim}, \bibinfo{person}{Stephen~R
  Marschner}, \bibinfo{person}{Marie-Paule Cani}, {and} \bibinfo{person}{Ming~C
  Lin}.} \bibinfo{year}{2007}\natexlab{}.
\newblock \showarticletitle{A survey on hair modeling: Styling, simulation, and
  rendering}.
\newblock \bibinfo{journal}{{\em IEEE Transactions on Visualization and
  Computer Graphics\/}} \bibinfo{volume}{13}, \bibinfo{number}{2}
  (\bibinfo{year}{2007}).
\newblock


\bibitem[\protect\citeauthoryear{Weng, Wang, Li, Chai, and Zhou}{Weng
  et~al\mbox{.}}{2013}]%
        {weng2013hair}
\bibfield{author}{\bibinfo{person}{Yanlin Weng}, \bibinfo{person}{Lvdi Wang},
  \bibinfo{person}{Xiao Li}, \bibinfo{person}{Menglei Chai}, {and}
  \bibinfo{person}{Kun Zhou}.} \bibinfo{year}{2013}\natexlab{}.
\newblock \showarticletitle{Hair interpolation for portrait morphing}. In
  \bibinfo{booktitle}{{\em Computer Graphics Forum}},
  Vol.~\bibinfo{volume}{32}. Wiley Online Library, \bibinfo{pages}{79--84}.
\newblock


\bibitem[\protect\citeauthoryear{Wu}{Wu}{2011}]%
        {wu2011visualsfm}
\bibfield{author}{\bibinfo{person}{Changchang Wu}.}
  \bibinfo{year}{2011}\natexlab{}.
\newblock \showarticletitle{VisualSFM: A visual structure from motion system}.
\newblock  (\bibinfo{year}{2011}).
\newblock


\bibitem[\protect\citeauthoryear{Xiong and {De la Torre}}{Xiong and {De la
  Torre}}{2013}]%
        {XiongD13}
\bibfield{author}{\bibinfo{person}{Xuehan Xiong} {and}
  \bibinfo{person}{Fernando {De la Torre}}.} \bibinfo{year}{2013}\natexlab{}.
\newblock \showarticletitle{Supervised Descent Method and its Applications to
  Face Alignment}. In \bibinfo{booktitle}{{\em IEEE Conference on Computer
  Vision and Pattern Recognition (CVPR)}}.
\newblock


\bibitem[\protect\citeauthoryear{Yacoob and Davis}{Yacoob and Davis}{2006}]%
        {yacoob2006detection}
\bibfield{author}{\bibinfo{person}{Yaser Yacoob} {and} \bibinfo{person}{Larry~S
  Davis}.} \bibinfo{year}{2006}\natexlab{}.
\newblock \showarticletitle{Detection and analysis of hair}.
\newblock \bibinfo{journal}{{\em IEEE transactions on pattern analysis and
  machine intelligence\/}} \bibinfo{volume}{28}, \bibinfo{number}{7}
  (\bibinfo{year}{2006}), \bibinfo{pages}{1164--1169}.
\newblock


\bibitem[\protect\citeauthoryear{Zhang, Chai, Wu, Yang, and Zhou}{Zhang
  et~al\mbox{.}}{2017}]%
        {zhang2017data}
\bibfield{author}{\bibinfo{person}{Meng Zhang}, \bibinfo{person}{Menglei Chai},
  \bibinfo{person}{Hongzhi Wu}, \bibinfo{person}{Hao Yang}, {and}
  \bibinfo{person}{Kun Zhou}.} \bibinfo{year}{2017}\natexlab{}.
\newblock \showarticletitle{A data-driven approach to four-view image-based
  hair modeling}.
\newblock \bibinfo{journal}{{\em ACM Transactions on Graphics (TOG)\/}}
  \bibinfo{volume}{36}, \bibinfo{number}{4} (\bibinfo{year}{2017}),
  \bibinfo{pages}{156}.
\newblock


\bibitem[\protect\citeauthoryear{Zheng, Jayasumana, Romera-Paredes, Vineet, Su,
  Du, Huang, and Torr}{Zheng et~al\mbox{.}}{2015}]%
        {zheng2015conditional}
\bibfield{author}{\bibinfo{person}{Shuai Zheng}, \bibinfo{person}{Sadeep
  Jayasumana}, \bibinfo{person}{Bernardino Romera-Paredes},
  \bibinfo{person}{Vibhav Vineet}, \bibinfo{person}{Zhizhong Su},
  \bibinfo{person}{Dalong Du}, \bibinfo{person}{Chang Huang}, {and}
  \bibinfo{person}{Philip~HS Torr}.} \bibinfo{year}{2015}\natexlab{}.
\newblock \showarticletitle{Conditional random fields as recurrent neural
  networks}. In \bibinfo{booktitle}{{\em Proceedings of the IEEE International
  Conference on Computer Vision}}. \bibinfo{pages}{1529--1537}.
\newblock


\bibitem[\protect\citeauthoryear{Zollh{\"o}fer, Nie{\ss}ner, Izadi, Rehmann,
  Zach, Fisher, Wu, Fitzgibbon, Loop, Theobalt, et~al\mbox{.}}{Zollh{\"o}fer
  et~al\mbox{.}}{2014}]%
        {zollhofer2014real}
\bibfield{author}{\bibinfo{person}{Michael Zollh{\"o}fer},
  \bibinfo{person}{Matthias Nie{\ss}ner}, \bibinfo{person}{Shahram Izadi},
  \bibinfo{person}{Christoph Rehmann}, \bibinfo{person}{Christopher Zach},
  \bibinfo{person}{Matthew Fisher}, \bibinfo{person}{Chenglei Wu},
  \bibinfo{person}{Andrew Fitzgibbon}, \bibinfo{person}{Charles Loop},
  \bibinfo{person}{Christian Theobalt}, {et~al\mbox{.}}}
  \bibinfo{year}{2014}\natexlab{}.
\newblock \showarticletitle{Real-time non-rigid reconstruction using an rgb-d
  camera}.
\newblock \bibinfo{journal}{{\em ACM Transactions on Graphics, TOG\/}}
  (\bibinfo{year}{2014}).
\newblock


\end{thebibliography}

\begin{appendices}
\label{appendix}
Hair segmentation is an important part of face parsing and full head reconstruction. \cite{yacoob2006detection} detected hair based on the position relationship between face and hair and a simple color model. \cite{wang2011novel} proposed a coarse-to-fine hair segmentation method that starts from a coarse candidate region and performs graph-cuts to segment the hair. A CNN-based face parsing method \cite{luo2012hierarchical} hierarchically combined several detectors to detect face components. \cite{liu2015multi,liu2017face}  proposed multi-objective learning frameworks that could parse facial components as well as hair regions, but this model requires facial landmarks as prior inputs and can only handle simple hairstyles. To allow  robust hair segmentation on various hairstyles, \cite{chai2016autohair} trained a deep network specifically for the hair regions. However, their method requires pre-alignment of the face to detect the hair region. In recent years, fully convolutional networks (FCN) \cite{long2015fully} have been widely used for pixel-level segmentation. We adopted the FCN model for robust hair segmentation and trained a network to segment hair regions across various poses.

We collected $15,977$ hair salon images from a hairstyle design website \cite{Hairbobo} \footnote{\url{http://www.hairbobo.com/faxingtupian}} and $3,923$ Internet images of celebrities from Google image search in various head poses and different hairstyles total of $19,900$. Each pixel in those images was labeled manually as hair or non-hair. To preserve continuity of the hair region, hair accessories were labeled as hair. The training images were not cropped nor aligned to increase robustness. 

\section{Hair Segmentation Classifier}
We trained the hair segmentation classifier using a fully convolutional network with $13,900$ images out of all the collected photos, and the remaining were used for testing. Specifically, the FCN-32s model \cite{long2015fully} was used, and it was fine-tuned with PASCAL VOC data from the ILSVRC-trained VGG-16 model. To only detect the hair category, we changed the output number of the last convolution layer to one and add a sigmoid layer to get scores between $0$ and $1$. The output score represents the probability that the pixel belongs to hair.
	In our implementation, we resized all our input images to $500 \times 500$. With FCN-32s, we downsampled the output with a factor of $32$. We later used bilinear interpolation to upsample the output heatmap to obtain the final segmentation result. We fine-tuned the pretrained FCN-32s with the following parameters: minibatch size $1$, learning rate $10^{-9}$, momentum $0.99$, and weight decay $0.0005$. We  froze all the layers except the last score layer in the first $100,000$ iterations. Then we fine-tuned all the layers in the next $100,000$ iterations.

We tested our segmenter on the $6,000$ test images. The hair segmentation network was implemented with Caffe \cite{jia2014caffe} and C++ and ran on a NVIDIA GTX 1080 GPU with an inference time of $150$ms for a $500 \times 500$ input image. The accuracy on the test images reached $0.9613$, and the IOU rate reached $0.8585$.

Using automatic hair segmentation (that works well across views, even back views) is the key to enabling a fully automatic hair modeling system. Our algorithm is capable of segmenting the hair region successfully in different views and head poses. However, it still fails to segment some hairstyles correctly when the hair color is too close to the background or when the image has a large motion blur.

We compared our classifier to two methods: \cite{liu2017face} and DeepLab \cite{chen2016deeplab} (which was compared to in \cite{chai2016autohair}, but \cite{chai2016autohair} does not provide code so we compared with DeepLab). We fine tuned and ran DeepLab on all of our test images and got a pixel accuracy of $0.8892$ and IOU rate of $0.7173$. For \cite{liu2017face}, we used the pre-trained model to run on only $41.3\%$ of our test images, since it requires pre-detection of the face and fails on back and side views. The pixel accuracy for the $41.3\%$ test images was $0.8462$, and IOU rate was $0.5573$.

\section{Hair Directional Classifier}
For training, the hair areas were manaully  divided into subregions based on their general growing trends. One of four directional labels was assigned in the labeling stage: $\left[ 0,0.5\pi\right)$, $\left[0.5\pi,\pi\right)$, $\left[\pi,1.5\pi\right)$, $\left[1.5\pi,2\pi\right)$. Hair accessories and hair occlusions were labeled as undetermined region, and background pixels were labeled as background. We trained a modified VGG16 network as proposed in \cite{chai2016autohair} on the hair regions to automatically predict the directional labels for each pixel, using a multi-class approach with 6 classes (4 directions, 1 background, 1 undetermined). 

To train our hair directional classifier, we cropped and extracted the hair region, resized each image to $256\times 256$ and downsampled $8$ times with a bilinear filter. The output result was then upsampled to the original image size with bilinear interpolation and then followed by a CRF for per-pixel labels. In the training stage, we utilized the same set of $13,900$ images and augmented the dataset to $20,000$ images by image rotation, translation, and mirroring. 

We implemented the classifier in the same environment as the segmenter and set the minibatch size to $32$ and the initial learning rate to $0.0001$ with exponential decay. Our network converged after $50$k steps.
The inference time was  $59$ms for a $256 \times 256$ input image. We ran the directional classifier on the same test set of $6,000$ images and got an accuracy of $0.9425$. 

Our classifier typically fails to generate a correct direction label for some small regions on the side of the face. However, in our pipeline, since we have video sequences, we can still get a correct direction from a different view. The availability of many views compensates for individual failure cases. 

\end{appendices}

\end{document}